%% file: neurips_main.tex
\documentclass{article}

\usepackage{multirow}
\usepackage[linesnumbered,algoruled,boxed,lined]{algorithm2e}
\SetKwProg{Fn}{Function}{}{end}\SetKwFunction{FRecurs}{FnRecursive}%

\SetCommentSty{mycommfont}
\usepackage{listings}
\SetKw{Continue}{continue}

\usepackage[final]{neurips_2025}

\usepackage[utf8]{inputenc} 
\usepackage[T1]{fontenc}    
\usepackage{hyperref}       
\usepackage{url}            
\usepackage{booktabs}       
\usepackage{amsfonts}       
\usepackage{nicefrac}       
\usepackage{microtype}      
\usepackage{xcolor}         
\usepackage{amsmath}
\usepackage{graphicx}
\usepackage{subcaption}

\usepackage{tikz, pgfplots, pgfplotstable, pgf-pie}
\usetikzlibrary{math}
\usetikzlibrary{patterns}
\usetikzlibrary{positioning}
\usepgfplotslibrary{groupplots}

\input{preamble}

\title{\ours: Contextualizing Embodied Agents via Scene-Graph Generation}

\author{%
  Jiani Huang \\
  University of Pennsylvania\\
  \texttt{jianih@seas.upenn.edu} \\
  \And
  Amish Sethi $^*$\\
  University of Pennsylvania \\
  \texttt{asethi04@seas.upenn.edu} \\
  \And
  Matthew Kuo $^*$\\
  University of Pennsylvania \\
  \texttt{mkuo@seas.upenn.edu} \\
  \And
  Mayank Keoliya \\
  University of Pennsylvania \\
  \texttt{mkeoliya@seas.upenn.edu} \\
  \And
  Neelay Velingker \\
  University of Pennsylvania \\
  \texttt{neelay@seas.upenn.edu} \\
  \And
  JungHo Jung \\
  University of Pennsylvania \\
  \texttt{j76jung@seas.upenn.edu} \\
  \And
  Ser-Nam Lim \\
  University of Central Florida \\
  \texttt{sernam@ucf.edu} \\
  \And
  Ziyang Li \\
  Johns Hopkins University \\
  \texttt{ziyang@cs.jhu.edu} \\
  \And
  Mayur Naik \\
  University of Pennsylvania \\
  \texttt{mhnaik@seas.upenn.edu}
}

\begin{document}
\def\thefootnote{*}\footnotetext{These authors contributed equally to this work}\def\thefootnote{\arabic{footnote}}

\maketitle
\input{sections/0_abstract}
\input{sections/1_introduction}

\input{sections/2_framework}

\input{sections/3_model}

\input{sections/4_experimental_setup}

\input{sections/5_evaluation}
\input{sections/6_related_works}
\input{sections/7_conclusion}

{
    \small
    \clearpage
    \bibliographystyle{plain}
    \bibliography{main}
}


\newpage
\input{sections/checklist}

\newpage
\input{sections/appendix}

\end{document}

%% file: preamble.tex
\newcommand{\ours}{{\sc ESCA}\xspace}

\newcommand{\ourmodel}{SGClip\xspace}
\newcommand{\ourdataset}{ESCA-Video-87K\xspace}

\definecolor{jianipink}{rgb}{0.859, 0.478, 0.576}

\definecolor{neelayblue}{rgb}{0.239, 0.427, 0.949}

\definecolor{amishred}{rgb}{0.851, 0.18, 0.075}

\definecolor{matthewpurple}{rgb}{0.675, 0.345, 0.788}

\definecolor{mayankgreen}{rgb}{0.443, 0.69, 0.11}

\definecolor{deepseek1}{RGB}{230,65,40}
\definecolor{deepseek2}{RGB}{204,65,37}
\definecolor{llamaOrange}{RGB}{234,145,56}
\definecolor{qwqPink}{RGB}{103,78,167}
\definecolor{neutralGray}{RGB}{100,100,100}
\definecolor{gpt1}{RGB}{74,134,232}
\definecolor{gpt2}{RGB}{49,130,189}

\definecolor{red}{RGB}{230,65,40}
\definecolor{orange}{RGB}{234,145,56}
\definecolor{pink}{RGB}{230,119,176}
\definecolor{purple}{RGB}{103,78,167}
\definecolor{green}{RGB}{50,168,82}
\definecolor{blue}{RGB}{79, 119, 176}
\definecolor{cyan}{RGB}{120, 211, 204}
\definecolor{yellow}{RGB}{226, 218, 107}

\newcommand{\secref}[1]{Section~\ref{#1}}

\newcommand{\algref}[1]{Algorithm~\ref{#1}}

\newcommand{\figref}[1]{Figure~\ref{#1}}
\newcommand{\tabref}[1]{Table~\ref{#1}}

\lstdefinelanguage{json}{
    basicstyle=\normalfont\ttfamily,
    showstringspaces=false,
    breaklines=true,
    frame=lines,
    literate=
        *{0}{{{\color{blue}0}}}{1}
         {1}{{{\color{blue}1}}}{1}
         {2}{{{\color{blue}2}}}{1}
         {3}{{{\color{blue}3}}}{1}
         {4}{{{\color{blue}4}}}{1}
         {5}{{{\color{blue}5}}}{1}
         {6}{{{\color{blue}6}}}{1}
         {7}{{{\color{blue}7}}}{1}
         {8}{{{\color{blue}8}}}{1}
         {9}{{{\color{blue}9}}}{1}
}

%% file: sections/0_abstract.tex
\begin{abstract}
Multi-modal large language models (MLLMs) are making rapid progress toward general-purpose embodied agents.
However, existing MLLMs do not reliably capture fine-grained links between low-level visual features and high-level textual semantics, leading to weak grounding and inaccurate perception.
To overcome this challenge, we propose \ours, a framework that contextualizes embodied agents by grounding their perception in spatial-temporal scene graphs.
At its core is \ourmodel, a novel, open-domain, promptable foundation model for generating scene graphs that is based on CLIP. 
\ourmodel is trained on 87K+ open-domain videos using a neurosymbolic pipeline that aligns automatically generated captions with scene graphs produced by the model itself, eliminating the need for human-labeled annotations.
We demonstrate that \ourmodel excels in both prompt-based inference and task-specific fine-tuning, achieving state-of-the-art results on scene graph generation and action localization benchmarks.
\ours with \ourmodel improves perception for embodied agents
 based on both open-source and commercial MLLMs, achieving state of-the-art performance across two embodied environments. 
Notably, \ours significantly reduces agent perception errors and enables open-source models to surpass proprietary baselines.
We release the source code for SGCLIP model training at \url{https://github.com/video-fm/LASER} and for the embodied agent at \url{https://github.com/video-fm/ESCA}.
\end{abstract}

%% file: sections/1_introduction.tex
\section{Introduction}
\label{sec:intro}

Recent advances in large-scale pretraining have enabled foundation models to assist with a wide range of tasks, from language and vision understanding~\citep{bai2023qwen, chen2024internvl, lin2023video, wu2023visual} to mathematical problem solving~\citep{zhang2024mathverse, didolkar2024metacognitive} and code generation~\citep{finnie2022robots, olausson2023linc, liu2024exploring}. 
However, it remains an open challenge to realize embodied agents that are capable of doing household chores, training alongside humans in physical activities, or providing care for the aging~\cite{kambhampati2024llmscantplanhelp, chen2024relyllmagentsdraft}. 
A critical first step toward this goal is equipping agents with fine-grained perception to ground abstract goals in physical interactions.

Despite progress, multi-modal large language models (MLLMs) still struggle to build spatially and temporally grounded world models.
Their inability to reliably ground visual features with spatial-temporal relations creates a disconnect between conceptual semantics and pixel-level observations~\citep{yang2024thinkingspacemultimodallarge, marsili2025visualagenticaispatial}. 
This lack of structured, fine-grained scene understanding severely limits their effectiveness in embodied environments. 
In fact, our empirical analysis shows that up to 69\% of agent failures stem from perception errors, highlighting the need for frameworks that can bridge this gap.

To address this challenge, we propose integrating structured scene graphs into the perception, reasoning, and planning pipelines of MLLM-based embodied agents. 
While prior work has explored enhancing MLLMs with external visual grounding modules, these approaches typically rely on open-domain object detection models such as Grounding DINO~\cite{liu2024groundingdinomarryingdino} and YOLO~\cite{hidayatullah2025yolo11}. 
However, these models are primarily designed for object identification and often overlook semantic attributes, inter-object relationships, and temporal consistency.

\begin{figure}
    \includegraphics[width=\linewidth]{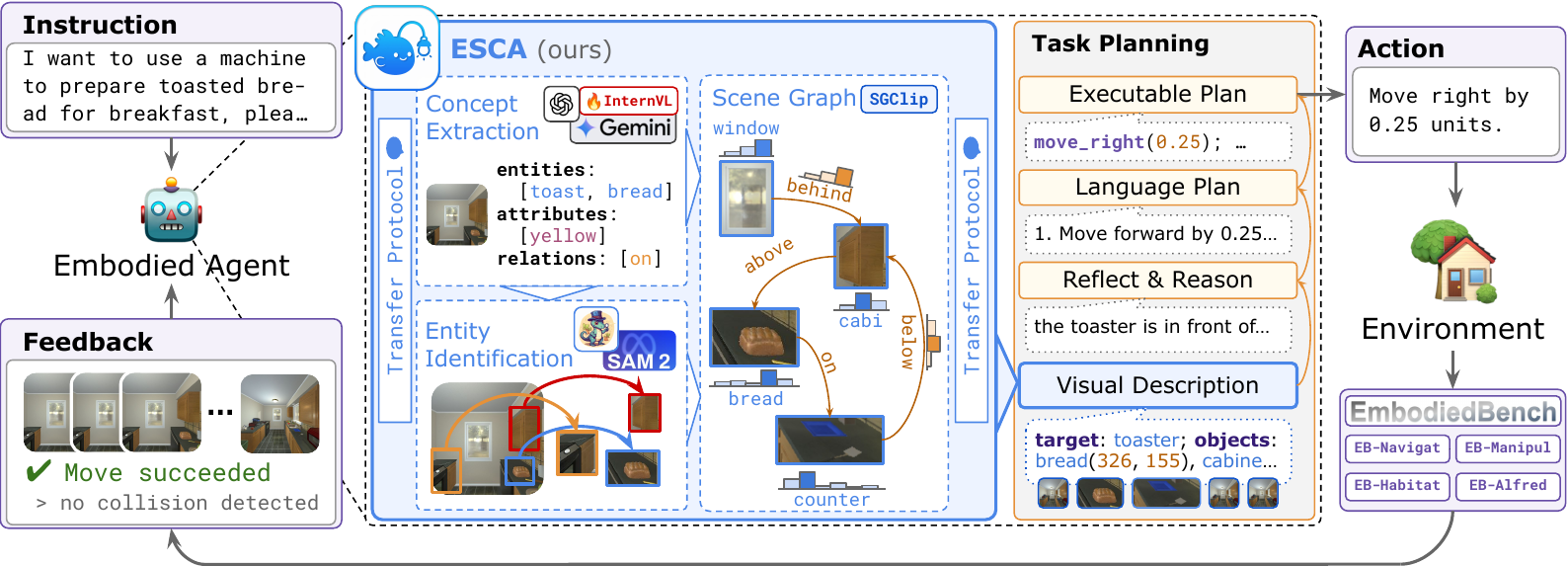}
    \caption{
        An overview of embodied agent pipeline augmented with ESCA. 
        In each cycle, the agent takes in an instruction and the environmental feedback and outputs a concrete executable action, through a sequence of perception, reasoning, and planning.
        The action is then executed and the environment will provide the next state.
        Notably, \ours contextualizes the task planner with grounded visual features represented as a scene graph.
    } 
\label{fig:banner}
\end{figure}

In this work, we introduce \textbf{ESCA} (\underline{\textbf{E}}mbodied and \underline{\textbf{S}}cene-Graph \underline{\textbf{C}}ontextualized \underline{\textbf{A}}gent), a framework designed to contextualize MLLMs through open-domain scene graph generation (\figref{fig:banner}). 
Much like the bioluminescent lure of a deep-sea anglerfish, which illuminates its surroundings to reveal otherwise hidden prey, \ours provides structured visual grounding that helps MLLMs make sense of complex and ambiguous sensory environments. 
A key feature of ESCA is selective grounding: rather than injecting full scene graphs, which may degrade performance, the MLLM first identifies the subset of objects, attributes, and relations most pertinent to the instruction, then determines the essential entities for task completion. 
This mechanism is supported by our transfer protocol, which performs probabilistic reasoning over object names, attributes, and spatial relations to construct prompts enriched with the most relevant scene elements. 
At its core is \ourmodel, a CLIP-based model that captures semantic visual features, including entity classes, physical attributes, actions, interactions, and inter-object relations. 


Through experiments on four challenging embodied environments, we demonstrate that \ours consistently improves the performance of all evaluated MLLMs, including both open-source and proprietary models. 
By providing structured and grounded scene graphs, \ours significantly reduces perception errors, laying the foundation for more reliable reasoning and planning. 
Beyond its integration with MLLM-based agents, we show that \ourmodel, when evaluated independently, exhibits strong zero-shot generalization, is promptable for task-specific scene understanding, and remains fine-tunable for downstream tasks such as action recognition.

In summary, our contributions are as follows:
(1) we present \ours, a general framework for contextualizing MLLM-based embodied agents through selective scene graph generation;
(2) we introduce the transfer protocol for enriching prompts with probabilistically inferred scene-specific information for diverse embodied benchmarks;
(3) we introduce \ourmodel, a generalizable and fine-grained scene graph generation model, along with \ourdataset, an MLLM annotated dataset;
(4) we conduct extensive evaluations demonstrating the effectiveness and versatility of \ours and \ourmodel across both embodied agent and scene understanding tasks.

%% file: sections/2_framework.tex
\section{ESCA: A Framework for Embodied Agents}
\label{sec:framework}

\subsection{Background}
\label{sec:framework-background}

\textbf{Embodied Environments.}
Recent research on embodied agents has been accelerated by the availability of simulated environments such as VisualAgentBench \citep{liu2024visualagentbenchlargemultimodalmodels} and EmbodiedBench \citep{yang2025embodiedbench}, which provide rich, multimodal task suites covering navigation, manipulation, and interaction across diverse scenarios. 
These benchmarks pose challenges that require agents to operate in both 
a) low-level action spaces such as continuous control signals over robot joints, and
b) high-level action spaces such as programmatic instructions and skills that abstract over low-level actions.

These tasks are typically modeled as Partially Observable Markov Decision Processes (POMDPs), represented as 7-tuple $(S, A, \Omega, \mathcal{T}, \mathcal{O}, L_\text{ins}, \mathcal{R})$, where 
$S$ is the unobservable state space, 
$A$ is the task-specific action space, 
$\Omega$ is the visual perception space where $I_t \in \Omega$ is an image frame at time $t$, 
$\mathcal{T} : S \times A \rightarrow S$ is the state transition dynamics, 
$\mathcal{O} : S \rightarrow \Omega$ relates the underlying state to the observations, 
$L_\text{ins}$ is the natural language instruction for the agent, and 
$\mathcal{R} : S \rightarrow \{0, 1\}$ is the reward function indicating whether the task has been completed.

\textbf{Embodied Agents and MLLM-Based Embodied Agents.}
A typical embodied agent interacts with the environment by maintaining a history of observations and actions:
$h_t = (I_0, a_0, I_1, \dots, a_{t-1}, I_t)$, where $a_t$ is the action taken by the agent at time $t$.
The agent selects the next action by conditioning on the instruction $L$ and history $h_t$ via a policy $\pi(a_t ~|~ L_\text{ins}, h_t)$.
An \textit{MLLM-based embodied agent} realizes such a policy by leveraging a multi-modal model that processes both:
a) imagery data $I_t$, and
b) textual data, including the instruction $L_\text{ins}$ and textual representations of actions $a \in A$.

Established by \cite{yang2025embodiedbench}, recent MLLM-based agent architectures often decompose the policy evaluation process into structured stages inspired by cognitive reasoning workflows (\figref{fig:banner}):
1) \textit{Visual Description}: extracting and summarizing visual inputs;
2) \textit{Reflection}: integrating observations with historical context to build situational awareness;
3) \textit{Reasoning}: inferring task-relevant insights based on the combined context;
4) \textit{Language Plan}: generating high-level plans or action sequences in natural language;
5) \textit{Executable Plan}: translating plans into concrete actions executable by the agent.

\textbf{Challenges of MLLM-Based Embodied Agents.}
Despite recent progress, MLLM-based embodied agents remain fragile in complex environments due to compounded errors across perception, reasoning, and planning~\cite{kambhampati2024llmscantplanhelp, chen2024relyllmagentsdraft}.
\textit{Perception errors} include hallucinated objects, misrecognized entities or actions, and incorrect spatial relationships. 
\textit{Reasoning errors} arise when agents fail to correctly infer spatial relations or recognize task termination states. 
These issues propagate into \textit{planning}, where agents may skip critical steps or generate invalid plans due to inaccurate state estimation.

\begin{figure}
    \includegraphics[width=\linewidth]{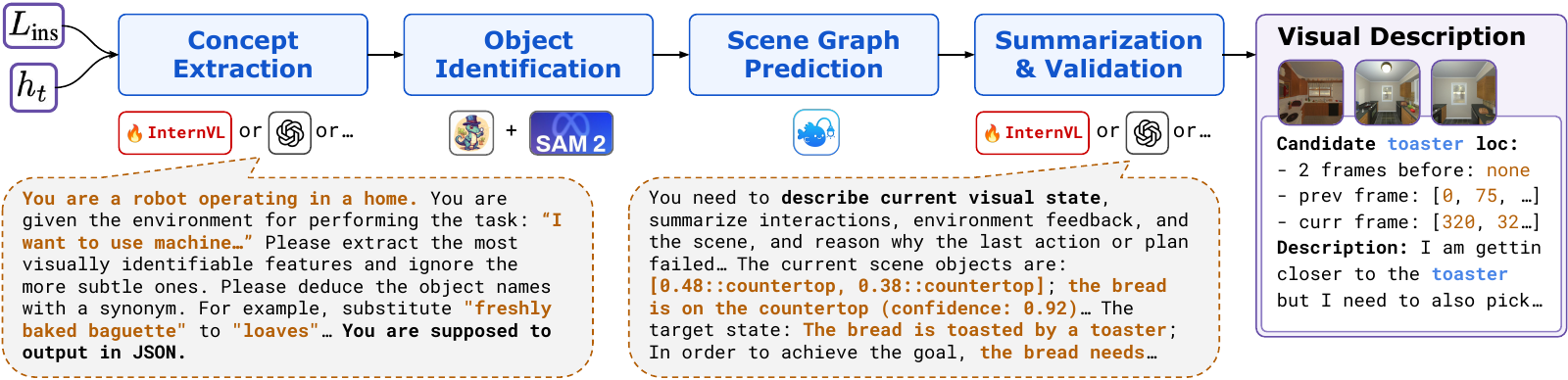}
    \caption{
        A detailed illustration of the visual description module, which involves concept extraction, object identification, scene graph prediction, and visual summarization.
        We also illustrate sample MLLM prompts used in a kitchen environment for the concept extraction and summarization steps.
    }
    \label{fig:visual-description}
\end{figure}

\subsection{The \ours Framework}
\label{sec:framework-esca}

At the core of \ours is a simple yet effective idea: 
contextualizing MLLM-based agents with grounded, structured scene-graph information to improve visual description. 
Specifically, given a language instruction $L_\text{ins}$ and interaction history $h_t$, the agent generates a multi-modal message sequence of images and text.
While existing approaches rely on end-to-end MLLMs to implicitly perform this step, \ours decomposes the process into four modular stages below (\figref{fig:visual-description}).

\textbf{Selective Concept Extraction.}
This module extracts concepts with MLLM guided by carefully designed prompts based on the instruction $L_\text{ins}$ and the history $h_t$.
Rather than producing only free-form natural descriptions, \ours requires the MLLM to explicitly extract structured concepts that are most relevant to the query, which can be classified as follows.
a) \textit{entity classes} such as (\texttt{car}, \texttt{knife}, \texttt{person}, \texttt{cup});
b) \textit{attributes} including physical properties (\texttt{red}, \texttt{small}, \texttt{broken}) and semantic states (\texttt{close-by}, \texttt{far}, \texttt{moving}, \texttt{sitting});
c) \textit{relations} covering spatial relations (\texttt{behind}, \texttt{above}) and interactions (\texttt{cutting}, \texttt{cooking}).

Concept extraction is guided by two signals:
1) the instruction $L_\text{ins}$, which highlights target entities, attributes, or relations the agent should focus on, and
2) the visual feedback $I_t$, which provides spatial context such as environmental structures (e.g., barriers, pathways) and dynamic interactions (e.g., objects being manipulated).
Overall, this MLLM-generated structured output provides a targeted concepts set $\bar{c} = \{c_1, c_2, \dots\}$ for subsequent object identification and scene graph generation.

\textbf{Object Identification.}
Given the extracted concepts and the agent's visual feedback, the second module grounds these concepts to specific image segments $\bar{\sigma} = \{\sigma_1, \sigma_2, \dots\}$ each represented by a bit-mask.
This step isolates visual elements in the frame that correspond to classified entities or attributes, enabling downstream semantic inference over localized visual units rather than raw pixels.

The object identification module is implemented using a multi-stage pipeline. 
First, Grounding DINO (GD)~\citep{liu2024groundingdinomarryingdino}, a vision-language detection model, takes the extracted concepts and the visual input to predict bounding boxes for the mentioned entities. 
These bounding boxes are further refined into precise segmentation masks using SAM2~\citep{ravi2024sam2segmentimages}, a segmentation model that produces pixel-accurate regions.
Leveraging the GD + SAM2 pipeline improves computational efficiency and offers better generalization to both attribute and relational concepts, as further detailed in the Appendix.

\textbf{Scene Graph Prediction.}
We then construct a scene graph (SG)~\citep{johnson2015image, xu2017scene, huang2025laser} by grounding the extracted concepts into structured visual elements.
Specifically, the SG contains two types of probabilistic facts:
1) unary facts of the form $p::c_i(\sigma_j)$, each describing an entity $\sigma_j$ with their classes, attributes, or states represented as $c_i$;
2) relational facts of the form $p::c_i(\sigma_j, \sigma_k)$, each representing that a relation or an interaction $c_i$ between a pair of grounded entities $\sigma_j$ and $\sigma_k$.
Notably, each predicted fact is associated with a confidence score denoted as $p$.
This probabilistic formulation allows the SG to capture uncertainty and distributions over possible scene interpretations.

Implementation-wise, we build \ourmodel, a CLIP-based model~\citep{sutskever2021clip} fine-tuned for open-domain scene graph prediction. 
The design of \ourmodel is guided by three aforementioned key desiderata. 
First, it supports open-domain concept coverage, enabling it to generalize beyond fixed taxonomies.
Second, it is adaptable to extracting diverse types of information, including entity classes, attributes, and inter-object relations. 
Third, it produces probabilistic predictions, allowing the model to capture uncertainties. 
We provide a more detailed discussion of \ourmodel and its training in \secref{sec:model}.

\textbf{Visual Summarization and Validation.}
This module distills these multi-modal signals into a list of messages to contextualize the agent's reasoner and planner. 
Concretely, the summarizer takes a prompt and is responsible for transforming the structured scene graph into natural descriptions, while also validating the consistency between the visual feedback and the underlying structured scene graph. 
While the summary is customizable via our transfer protocol, the set of messages include
1) the current view (and potentially historic ones) augmented with visualized bounding boxes served as markers,
2) image segments corresponding to key entities,
3) the textual description of scene graph and segments, and
4) an analysis of history actions and how they caused the current scene.

\subsection{Transfer Protocol}
\label{sec:framework-implementation}

To enable \ours to generalize across different downstream tasks, we define a general transfer protocol based on the customization of two prompt templates, positioned at the entry and exit points of the entire visual description module. 
The goal of this unified transfer protocol is to maximize adaptability of \ours across tasks with diverse planning strategies, action spaces, and reasoning requirements, while maintaining a consistent interface. 

Specifically, the first prompt, the \textit{Concept Extraction Prompt}, specifies the required JSON output format and indicates, or enumerates if possible, task-specific concepts that the agent should focus on.
The second prompt, the \textit{Visual Summarization Prompt}, guides the model to produce a contextualized summary that integrates both grounded image segments and task-specific textual elements, such as target objects, desired states, and environmental constraints.
Together, these prompts provide a principled way to adapt \ours to diverse embodied AI tasks without retraining the core system.
We present a case study of the transfer protocol in \figref{fig:visual-description} and provide further details in \secref{sec:experimental-setup}.

%% file: sections/3_model.tex
\section{SGClip Model}
\label{sec:model}

To enable the generation of spatial-temporal scene graphs in an open-domain setting, especially the embodied environments, we develop \ourmodel, a CLIP-based foundation model~\citep{sutskever2021clip} for structured scene understanding. 
\ourmodel is designed to operate in an open-domain fashion, recognizing a wide and extensible set of concepts. 
Secondly, it must adapt to different types of concepts, while be capable of generalizing to unseen visual and textual domains. 
Finally, it must produce probabilistic predictions to capture uncertainty.

To meet these goals, \ourmodel builds on CLIP's vision-language architecture, which naturally supports joint reasoning over images and textual phrases. 
However, deploying CLIP directly is insufficient, as it lacks specialization for structured scene graph prediction. 
To bridge this gap, we fine-tune \ourmodel to balance adaptability and generalizability. 
In this section, we describe 
1) how to handle different types of concepts through inference time adaptation (\secref{sec:model-inference}),
2) how to overcome the lack of data by collecting a model-driven self-supervision dataset (\secref{sec:model-dataset}), and
3) how to learn without relying on human annotations via a self-supervised neurosymbolic learning pipeline (\secref{sec:model-learning}).


\begin{figure}
    \begin{subfigure}{0.25\linewidth}
        \includegraphics[width=\linewidth]{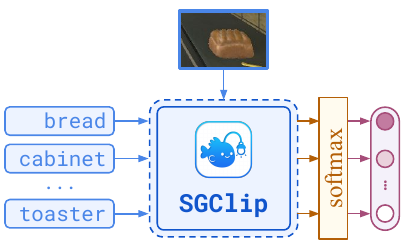}
        \caption{Entity classes}
    \end{subfigure}
    \hfill
    \begin{subfigure}{0.29\linewidth}
        \includegraphics[width=\linewidth]{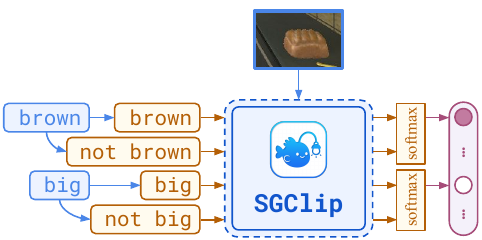}
        \caption{Attributes}
    \end{subfigure}
    \hfill
    \begin{subfigure}{0.44\linewidth}
        \includegraphics[width=\linewidth]{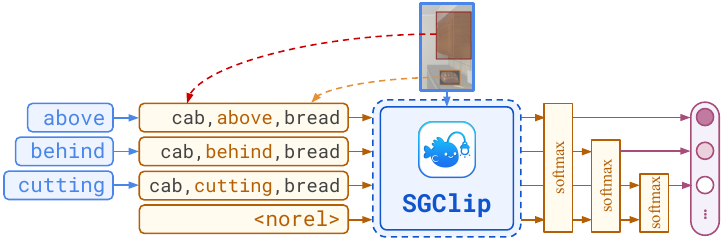}
        \caption{Binary relations}
    \end{subfigure}
    \caption{
        Illustration of the inference modes of \ourmodel for three types of concepts: entity classes, attributes, and binary relations.
        While the model stays the same, the three inference modes perform different pre- and post-processing for a more accurate semantic estimation of probabilities.
    }
    \label{fig:sgclip-inference}
\end{figure}

\subsection{Model Architecture and Inference Time Adaptation}
\label{sec:model-inference}

At its core, \ourmodel (Scene Graph CLIP) is a single CLIP-based model designed to score concept relevance within an image. 
Formally, it operates as $\text{SGClip}(\sigma, \bar{c}) \in \mathbb{R}^{|\bar{c}|}$, where $\sigma$ is the input image and $\bar{c}$
is the a set of candidate concepts, producing a logit score for each concept that reflects the model's confidence in its presence.
\ourmodel supports three distinct inference modes to handle different types of concepts: entity classes $\bar{c}_\text{class}$, attributes $\bar{c}_\text{attr}$, and binary relations $\bar{c}_\text{rela}$. 
Illustrated in \figref{fig:sgclip-inference}, each mode requires a specialized formulation of the input concepts and scoring process, effectively allowing \ourmodel to operate flexibly across these concept types:

\textbf{Entity classes.}
In this setting, \ourmodel is used to identify the most likely entity class presented in an image segment.
Let $\bar{c}_{\text{class}}$ denote the list of candidate entity classes. 
Since an entity is typically assumed to belong to a single class, we apply softmax normalization over the logit scores produced by \ourmodel for these candidates: $\text{softmax}(\text{SGClip}(\sigma, \bar{c}_\text{class}))$. 

\textbf{Attributes.}
To estimate the likelihood that a specific segment $\sigma$ possesses a particular attribute $c$, we construct a binary contrast between the attribute and its negation by evaluating $\text{softmax}(\text{SGClip}(\sigma, \{c, \neg c\}))$, where $\neg c$ denotes the negated textual phrase (e.g., ``not red'' for the attribute ``red''). 
The first element of the resulting probability distribution corresponds to the model's estimated likelihood. 
To improve computational efficiency, we perform batched evaluation by merging all attribute-contradiction pairs into a single concept set $\bar{c}_\text{attr}^* = \bar{c}_\text{attr} \cup \{\neg c ~|~ c \in \bar{c}_\text{attr}\}$. 

\textbf{Binary relations.}
For binary relation prediction, the goal is to determine whether a relation $c$ holds between two segments $\sigma_i$ and $\sigma_j$. 
To do this, we first compute a bounding region $\sigma_{ij}^*$ that tightly encloses both segments. 
Within this region, we apply distinct color tinting to $\sigma_i$ and $\sigma_j$ to indicate their directional roles (subject and object). 
To provide additional relational context, especially for interactions like ``\texttt{cutting},'' we augment the relation phrase by including the predicted entity classes of the subject and object, generating ``(\texttt{robot}, \texttt{cutting}, \texttt{cabbage)}''. 
Relation predictions are thus conditioned on the classes of both the subject and the object.
Specifically, for each segment $\sigma_i$, we compute its most likely class $\nu_i$ by selecting the top prediction from the entity class:
\[
\nu_i = c_{\text{class}\,u}, \quad \text{where } u = \text{argmax}_{u \in 1 \dots |\bar{c}_{\text{class}}|}\text{SGClip}(\sigma_i, \bar{c}_{\text{class}})_u.
\]
We then form the augmented relation phrase as $(\nu_i, c, \nu_j)$.
Similar to attribute prediction, we contrast the candidate relation with a special token \texttt{<norel>} denoting ``no relation,'' and compute $\text{softmax}(\text{SGClip}(\sigma_{ij}^*, \{(\nu_i, c, \nu_j), \texttt{<norel>}\}))$, to obtain the probability of whether the relation $c$ holds between object $i$ and $j$.
In practice, this process is batched over all segment pairs and relation concepts to maximize efficiency: $\bar{c}_{\text{rela}}^* = \{(\nu_i, c, \nu_j) ~|~ i, j \in 1\dots|\bar{\sigma}|, c \in \bar{c}_\text{rela}\} \cup \{\texttt{<norel>}\}$.

\begin{figure}
\includegraphics[width=\linewidth]{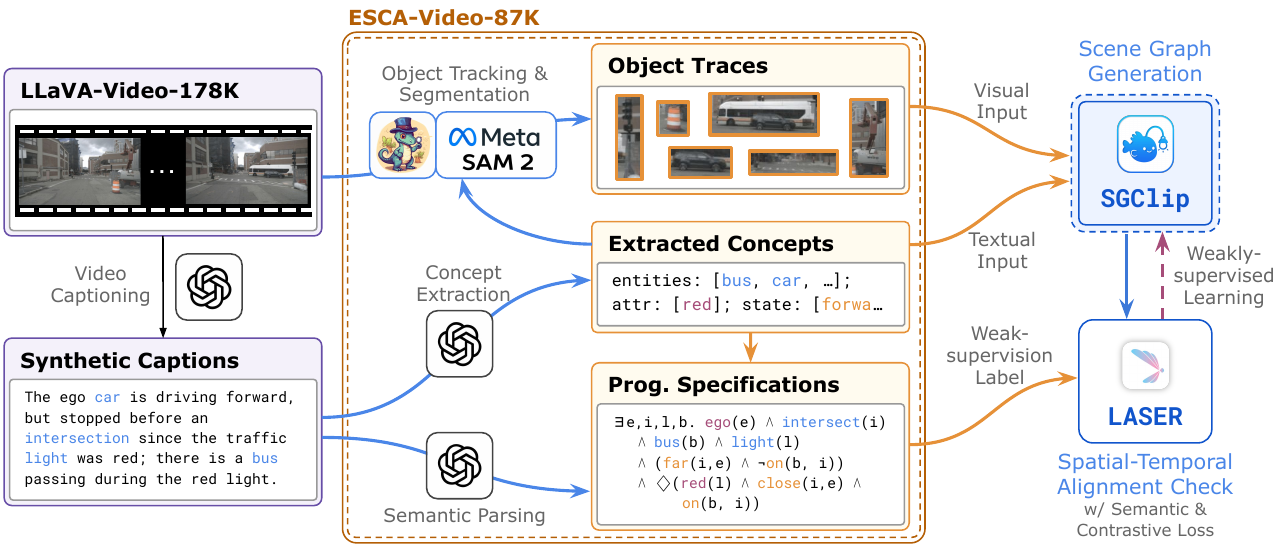}
\caption{
    Illustration of the construction of \ourdataset dataset and the model-driven self-supervised fine-tuning pipeline of our \ourmodel model.
    In addition to videos and their natural language captions, \ourdataset includes object traces, open-domain concepts, and programmatic specifications for 87K video-caption pairs.
    The dataset is then used to train \ourmodel via LASER~\citep{huang2025laser}, a neurosymbolic learning procedure based on spatial-temporal alignment.
}
\label{fig:sgclip_pipeline}
\end{figure}

\subsection{\ourdataset Dataset}
\label{sec:model-dataset}

We adopt the neurosymbolic weak-supervision pipeline introduced in LASER~\citep{huang2025laser}, which enables learning fine-grained STSGs from \textit{weak supervision signals} derived from spatial-temporal programmatic specifications, eliminating the need for costly manual annotations. 
While the details of this learning pipeline are provided in \secref{sec:model-learning}, we begin by introducing the \ourdataset dataset, the dataset we curate and use to train \ourmodel.

The \ourdataset dataset is constructed from the publicly available LLaVA-Video-178K dataset~\citep{zhang2024videoinstructiontuningsynthetic}, and consists of 87K short video clips, each paired with natural language captions generated by GPT-4~\citep{hurst2024gpt}.
As illustrated in \figref{fig:sgclip_pipeline}, these captions are first processed to extract relevant concepts which are then fed into GD~\citep{liu2024groundingdinomarryingdino} and SAM2~\citep{ravi2024sam2segmentimages} to obtain object traces, which are sequences of object segmentations that evolve across multiple video frames. 
In addition, these concepts are also used to assist in generating spatial-temporal programmatic specifications, expressed in a linear temporal logic-based language. 
To construct these specifications, we develop a semantic parsing pipeline, again leveraging GPT-4, which converts high-level captions into structured temporal statements. 
These specifications formally describe how the semantics of object traces evolve, capturing temporal relations using operators such as ``until'', ``finally'', or ``always''.

In summary, each data point in \ourdataset is represented as a 5-tuple $(\bar{I}, L_{\text{cap}}, \Sigma, \bar{c}, \phi)$, where 
$\bar{I} = \{I_1, I_2, \dots\}$ is the video, 
$L_{\text{cap}}$ is the associated natural language caption, 
$\Sigma = \{\bar{\sigma}_1, \bar{\sigma}_2, \dots\}$ is the set of object traces, 
$\bar{c} = \{c_1, c_2, \dots\}$ is the set of extracted concepts, and 
$\phi$ is the spatial-temporal programmatic specification. 
This rich, multi-level annotation enables training models like \ourmodel without requiring manual scene graph labeling. 
We defer additional details about the dataset construction process and data statistics to the Appendix.

\subsection{Neurosymbolic Learning Pipeline}
\label{sec:model-learning}

Given the \ourdataset, our goal is to fine-tune the \ourmodel model using the provided object traces, concepts, and spatial-temporal programmatic specifications. 
This is achieved by aligning the scene graphs generated by \ourmodel with the expected specifications~\citep{huang2025laser}, where the degree of alignment serves as the learning signal (\figref{fig:sgclip_pipeline}). 
To perform this alignment in a differentiable manner, we leverage the Scallop programming language~\citep{li2023scalloplanguageneurosymbolicprogramming}, enabling symbolic alignment checks to be integrated into end-to-end gradient-based learning. 

Specifically, the alignment loss computation mirrors the inference-time adaptation procedure described in \secref{sec:model-inference}, where different types of concepts are processed differently but unified under the same model.
The pipeline is further enhanced with \textit{semantic losses}, derived from evaluating common-sense and temporal constraint satisfaction, as well as a \textit{contrastive loss} that encourages the model to distinguish between matched and unmatched scene graph–specification pairs. 
Additional details of the training process are provided in the Appendix.

%% file: sections/4_experimental_setup.tex
\section{Embodied Environments and Transfer Protocol Setup}
\label{sec:experimental-setup}

We evaluate our approach on EmbodiedBench~\citep{li2024embodied}, a benchmark suite designed to assess MLLM-based embodied agents. 
We focus on two environments: EB-Navigation and EB-Manipulation, each of which requires different levels of perception, reasoning, and control, and may benefit from a contextualized visual description.
To adapt \ours to these tasks, we apply our transfer protocol by designing two specialized prompts for each environment. 
Full prompt templates are provided in the Appendix; here, we summarize the core challenges and how \ours addresses them.

\textbf{EB-Navigation} 
is built on AI2-THOR~\citep{ali2017ai2thor} and focuses on visual navigation tasks where the agent must locate target objects based on language instructions, such as ``navigate to the laptop.'' 
The agent relies solely on egocentric visual input and textual feedback, navigating through a space using eight low-level movement and rotation actions.
Existing models often fail by generating correct high-level plans but producing incorrect low-level actions due to poor spatial grounding from an egocentric perspective. 
\ours addresses this by generating accurate scene graphs that capture spatial relations and object positions. 
With the scene graphs, we design prompts that incorporate \textit{numerical bounding box data} and \textit{temporal movement cues}, helping the agent to localize targets more precisely.

\textbf{EB-Manipulation} 
extends VLMBench~\citep{zheng2022vlmbenchcompositionalbenchmarkvisionandlanguage} for evaluating low-level robotic manipulation. 
The agent controls a 7-DoF robotic arm using discretized action spaces, with additional signals such as YOLO bounding boxes~\citep{hidayatullah2025yolo11} and global 3D object pose estimates to assist manipulation.
Existing models struggle to ground object concepts into actionable spatial representations, leading to perception failures that disrupt downstream planning and control. 
\ours improves this by grounding target features into precise visual segments, enabling prompts that describe object attributes, semantic relations, and \textit{3D spatial coordinates}, giving the agent more reliable geometric context.

\textbf{EB-Habitat} builds upon Language Rearrangement task~\citep{szot2024large}, simulated via Habitat 2.0~\citep{szot2021habitat}, and primarily evaluates high-level task decomposition and planning capabilities.
The action space is limited to atomic high-level actions, such as \texttt{navigate}/\texttt{pick}/\texttt{place}/\texttt{open}/\texttt{close}, from which the agent is instructed to complete tasks such as ``Find a toy airplane and move it to the right counter''. 
Common failure cases of existing models include invalid actions arising from failure to identify and remember object displacements in the scene. 
Our model improves this by managing scene graphs of the current and desired state of the target object, which improves awareness of task progression and limits the number of focus objects.


\textbf{EB-Alfred} is based on the ALFRED dataset \cite{shridhar2020alfred} and AI2-THOR \cite{ali2017ai2thor}. It evaluates agents on high-level household tasks involving eight skill types like ``pick up'' or ``turn off.'' 
The agent receives egocentric observations and textual feedback on action validity, performing actions on objects.
While agents receive egocentric observations and action feedback, existing models tend to repeat the same mistakes because they fail to reflect on how past actions influence the current state. 
\ours addresses this by generating scene graphs that describe both the current and target states \textit{symbolically}, enabling prompts that support \textit{causal reasoning}. 
This allows the agent to recognize how previous actions led to the current situation and to deduce the necessary state changes to achieve the task goal.

%% file: sections/5_evaluation.tex
\section{Empirical Evaluation}
\label{sec:evaluation}

\input{figures/performance-charts}

Our experiments are designed to address two key research questions:
(1) How effectively does \ours, together with \ourmodel, improve embodied agent performance through structured scene graph generation? and
(2) How generalizable and adaptable is \ourmodel when evaluated independently on open-domain, zero-shot, and downstream transfer tasks?
We now detail our experimental setup and present empirical results addressing both questions.


\textbf{Experimental Setup.}
We evaluate \ours in EB-Navigation, EB-Manipulation, EB-Habitat, and EB-Alfred, four environments that demand fine-grained perception to support both low-level and high-level control. 
To assess the general applicability of \ours, we integrate it with four diverse MLLMs: 
InternVL-2.5-38B-MPO~\citep{chen2025internvl2_5mpo} , 
Qwen2.5-VL-72B-Ins~\citep{bai2025qwen25vltechnicalreport}, 
Gemini-2.0-flash~\citep{Sundar2024gemini2_0}, and
GPT-4o~\citep{hurst2024gpt}.
For each MLLM experiment, we use the model for both the concept extraction and visual summarization steps (\figref{fig:visual-description}).
To further benchmark \ours's impact, we compare to performance of MLLMs augmented with existing visual grounding modules, including Grounding DINO~\citep{liu2024groundingdinomarryingdino} and Ultralytics-YOLO11~\citep{hidayatullah2025yolo11}. 

For evaluating \ourmodel independently, we consider out-of-domain scene graph benchmarks, including OpenPVSG~\citep{yang2023panopticvideoscenegraph}, Action Genome~\citep{ji2020action}, and VidVRD~\citep{shang2017video}, comparing \ourmodel against strong baselines such as CLIP~\citep{sutskever2021clip}, InternVL-6B~\citep{chen2024internvl}, BIKE~\citep{wu2023bidirectional}, and Text4Vis~\citep{wu2023text4vis}. 
To assess \ourmodel's downstream adaptability beyond structured scene graph prediction, we further test the fine-tunability on the ActivityNet action recognition dataset by applying a transfer protocol. 


\input{figures/esca-analysis}

\textbf{\ours for Embodied Agents.}
As shown in \figref{fig:embodied-bench-performance}, \ours-augmented MLLMs consistently outperform their non-contextualized baselines across both EB-Navigation and EB-Manipulation. 
Remarkably, on EB-Navigation, even the open-source InternVL-2.5, when augmented with \ours, surpasses the base performance of the proprietary GPT-4o model. 
While integrating Grounding DINO or YOLO improves baseline models, \ours provides additional, substantial gains. 
For example, Gemini-2.0 with \ours achieves over 10\% improvement, while GPT-4o, already boosted by YOLO, still benefits from an additional 6\% performance gain on EB-Manipulation. 

Through qualitative analysis of end-to-end agent behaviors, we observe that \ours consistently improves the agent's perceptual grounding, leading to more effective task execution. 
As illustrated in \figref{fig:qualitative-example}, an agent powered by InternVL with \ours successfully identifies the kettle early in the episode and navigates directly toward it. 
In contrast, the base InternVL model fails to recognize the target and ultimately collapses onto the wall.
This observation is further supported by the error decomposition analysis shown in \figref{fig:error-decomposition}, where we find that \ours reduces the overall perception error rate from 69\% to 30\%. 
We provide additional detailed experimental results in the Appendix.

\textbf{Generalizability and Adaptability of \ourmodel.}
Evaluating \ourmodel's zero-shot generalization, \figref{fig:sgclip-zs-gen} shows that \ourmodel trained on \ourdataset consistently outperforms CLIP on OpenPVSG, Action Genome, and VidVRD, demonstrating strong out-of-domain robustness. 
Further, \ourmodel shows strong adaptability, achieving notable improvements when fine-tuned on VidVRD (details provided in the Appendix). 
Beyond scene graph tasks, \figref{fig:action-recog-finetunability} highlights \ourmodel's downstream transferability to action recognition on ActivityNet. 
Fine-tuned with only 1\% of the training data, \ourmodel outperforms state-of-the-art zero-shot video recognition baselines. 
With 5\% of the data (approximately 800 videos), \ourmodel achieves 92.10\% accuracy, approaching the performance of InternVideo2-6B with end-to-end finetuning on the ActivityNet dataset.

\input{figures/adaptation_experiments}

%% file: figures/performance-charts.tex
\begin{figure}
    \footnotesize

    \pgfplotscreateplotcyclelist{customcolors}{
        {fill=deepseek1!30!white,draw=deepseek1!50!black},      
        {fill=llamaOrange!30!white,draw=llamaOrange!50!black},    
        {fill=gpt1!80!white,draw=gpt1!50!black},    
        {fill=gpt2!50!white,draw=gpt2!50!black},    
        {fill=qwqPink!50!white,draw=qwqPink!50!black},      
    }
    
    \begin{subfigure}{0.48\linewidth}
        \centering
        \begin{tikzpicture}
            \begin{axis}[
                ybar,
                width=\linewidth,
                height=4.2cm,
                enlarge x limits=0.2,
                ylabel={Success Rate (\%)},
                y label style={at={(axis description cs:0.08,.5)}},
                xtick style={
                    draw=none, 
                },
                symbolic x coords={intern-vl-2.5-38b-mpo, qwen2.5-vl-72b-ins, gemini-2.0-flash, gpt-4o},
                xticklabels={IVL2.5, Qwen2.5, Gem2.0, GPT4o},
                xtick={intern-vl-2.5-38b-mpo, qwen2.5-vl-72b-ins, gemini-2.0-flash, gpt-4o},
                x tick label style={
                    yshift=1.5mm,
                    font=\footnotesize,
                },
                ymin=18, ymax=62,
                ymajorgrids=true,
                ytick={20, 30, 40, 50, 60},
                legend style={
                    at={(0.5,1)},
                    anchor=south,
                    legend columns=3,
                    column sep=3px,  
                    draw=none,
                    fill=none,
                    font=\footnotesize
                },
                cycle list name=customcolors,
                x=1.4cm,                  
                bar width=0.30cm,         
            ]
                \pgfplotstableread[col sep=comma]{data/eb_nav.csv}\datatable
        
                \addplot+[
                    bar shift=-0.3cm,
                ] table [x=model, y=base] {\datatable}; \addlegendentry{Base}
                \addplot+[
                    bar shift=0
                ] table [x=model, y=w_gd] {\datatable}; \addlegendentry{+ GD}
                \addplot+[
                    bar shift=0.3cm,
                    nodes near coords,
                    nodes near coords style={
                        font=\scriptsize, 
                        color=gpt1!60!black
                    },
                ] table [x=model, y=w_esca] {\datatable}; \addlegendentry{+ ESCA}
            \end{axis}
        \end{tikzpicture}
        \caption{EB-Navigation}
        \label{fig:perf-chart-eb-navigation}
    \end{subfigure}
    \hfill
    \begin{subfigure}{0.48\linewidth}
        \centering
        \begin{tikzpicture}
            \begin{axis}[
                ybar,
                width=\linewidth,
                height=4.2cm,
                enlarge x limits=0.2,
                xtick style={
                    draw=none, 
                },
                symbolic x coords={intern-vl-2.5-38b-mpo, qwen2.5-vl-72b-ins, gemini-2.0-flash, gpt-4o},
                xticklabels={IVL2.5, Qwen2.5, Gem2.0, GPT4o},
                xtick={intern-vl-2.5-38b-mpo, qwen2.5-vl-72b-ins, gemini-2.0-flash, gpt-4o},
                x tick label style={
                    yshift=1.5mm,
                    font=\footnotesize,
                },
                ymin=0, ymax=42,
                ymajorgrids=true,
                ytick={0, 10, 20, 30, 40},
                legend style={
                    at={(0.5,1)},
                    anchor=south,
                    legend columns=3,
                    column sep=3px,  
                    draw=none,
                    fill=none,
                    font=\footnotesize
                },
                cycle list name=customcolors,
                x=1.4cm,                  
                bar width=0.30cm,         
            ]
                \pgfplotstableread[col sep=comma]{data/eb_mani.csv}\datatable
        
                \addplot+[
                    bar shift=-0.3cm,
                ] table [x=model, y=base] {\datatable}; \addlegendentry{Base}
                \addplot+[
                    bar shift=0
                ] table [x=model, y=w_yolo] {\datatable}; \addlegendentry{+ YOLO}
                \addplot+[
                    bar shift=0.3cm,
                    nodes near coords,
                    nodes near coords style={
                        font=\scriptsize, 
                        color=gpt1!60!black
                    },
                ] table [x=model, y=w_esca] {\datatable}; \addlegendentry{+ ESCA}
            \end{axis}
        \end{tikzpicture}
        \caption{EB-Manipulation}
        \label{fig:perf-chart-eb-manipulation}
    \end{subfigure}

    \vspace{10px}

    \begin{subfigure}{0.48\linewidth}
        \centering
        \begin{tikzpicture}
            \begin{axis}[
                ybar,
                width=\linewidth,
                height=4.2cm,
                enlarge x limits=0.2,
                ylabel={Success Rate (\%)},
                y label style={at={(axis description cs:0.08,.5)}},
                xtick style={
                    draw=none, 
                },
                symbolic x coords={intern-vl-2.5-38b-mpo, qwen2.5-vl-72b-ins, gemini-2.0-flash, gpt-4o},
                xticklabels={IVL2.5, Qwen2.5, Gem2.0, GPT4o},
                xtick={intern-vl-2.5-38b-mpo, qwen2.5-vl-72b-ins, gemini-2.0-flash, gpt-4o},
                x tick label style={
                    yshift=1.5mm,
                    font=\footnotesize,
                },
                ymin=18, ymax=72,
                ymajorgrids=true,
                ytick={20, 30, 40, 50, 60, 70},
                legend style={
                    at={(0.5,1)},
                    anchor=south,
                    legend columns=3,
                    column sep=3px,  
                    draw=none,
                    fill=none,
                    font=\footnotesize
                },
                cycle list name=customcolors,
                x=1.4cm,                  
                bar width=0.30cm,         
            ]
                \pgfplotstableread[col sep=comma]{data/eb_habitat.csv}\datatable
        
                \addplot+[
                    bar shift=-0.3cm,
                ] table [x=model, y=base] {\datatable}; \addlegendentry{Base}
                \addplot+[
                    bar shift=0
                ] table [x=model, y=w_gd] {\datatable}; \addlegendentry{+ GD}
                \addplot+[
                    bar shift=0.3cm,
                    nodes near coords,
                    nodes near coords style={
                        font=\scriptsize, 
                        color=gpt1!60!black
                    },
                ] table [x=model, y=w_esca] {\datatable}; \addlegendentry{+ ESCA}
            \end{axis}
        \end{tikzpicture}
        \caption{EB-Habitat}
        \label{fig:perf-chart-eb-habitat}
    \end{subfigure}
    \hfill
    \begin{subfigure}{0.48\linewidth}
        \centering
        \begin{tikzpicture}
            \begin{axis}[
                ybar,
                width=\linewidth,
                height=4.2cm,
                enlarge x limits=0.2,
                y label style={at={(axis description cs:0.08,.5)}},
                xtick style={
                    draw=none, 
                },
                symbolic x coords={intern-vl-2.5-38b-mpo, qwen2.5-vl-72b-ins, gemini-2.0-flash, gpt-4o},
                xticklabels={IVL2.5, Qwen2.5, Gem2.0, GPT4o},
                xtick={intern-vl-2.5-38b-mpo, qwen2.5-vl-72b-ins, gemini-2.0-flash, gpt-4o},
                x tick label style={
                    yshift=1.5mm,
                    font=\footnotesize,
                },
                ymin=18, ymax=62,
                ymajorgrids=true,
                ytick={20, 30, 40, 50, 60},
                legend style={
                    at={(0.5,1)},
                    anchor=south,
                    legend columns=3,
                    column sep=3px,  
                    draw=none,
                    fill=none,
                    font=\footnotesize
                },
                cycle list name=customcolors,
                x=1.4cm,                  
                bar width=0.30cm,         
            ]
                \pgfplotstableread[col sep=comma]{data/eb_alfred.csv}\datatable
        
                \addplot+[
                    bar shift=-0.3cm,
                ] table [x=model, y=base] {\datatable}; \addlegendentry{Base}
                \addplot+[
                    bar shift=0
                ] table [x=model, y=w_gd] {\datatable}; \addlegendentry{+ GD}
                \addplot+[
                    bar shift=0.3cm,
                    nodes near coords,
                    nodes near coords style={
                        font=\scriptsize, 
                        color=gpt1!60!black
                    },
                ] table [x=model, y=w_esca] {\datatable}; \addlegendentry{+ ESCA}
            \end{axis}
        \end{tikzpicture}
        \caption{EB-Alfred}
        \label{fig:perf-chart-eb-alfred}
    \end{subfigure}

    \caption{
        The overall performance on EB-Navigation and EB-Manipulation environments.
        We show the performance of four base models,
        InternVL-2.5-38B-MPO (IVL2.5), 
        Genimi-2.0-flash (Gem2.0),
        Qwen2.5-VL-72B-Instruct (Qwen2.5), and
        GPT-4o (GPT4o),
        as well as their performance when accompanied with baseline visual grounding modules such as Grounding DINO (GD) or YOLO.
        With \ours and \ourmodel, all models consistently outperform the baselines.
    }
    \label{fig:embodied-bench-performance}
\end{figure}
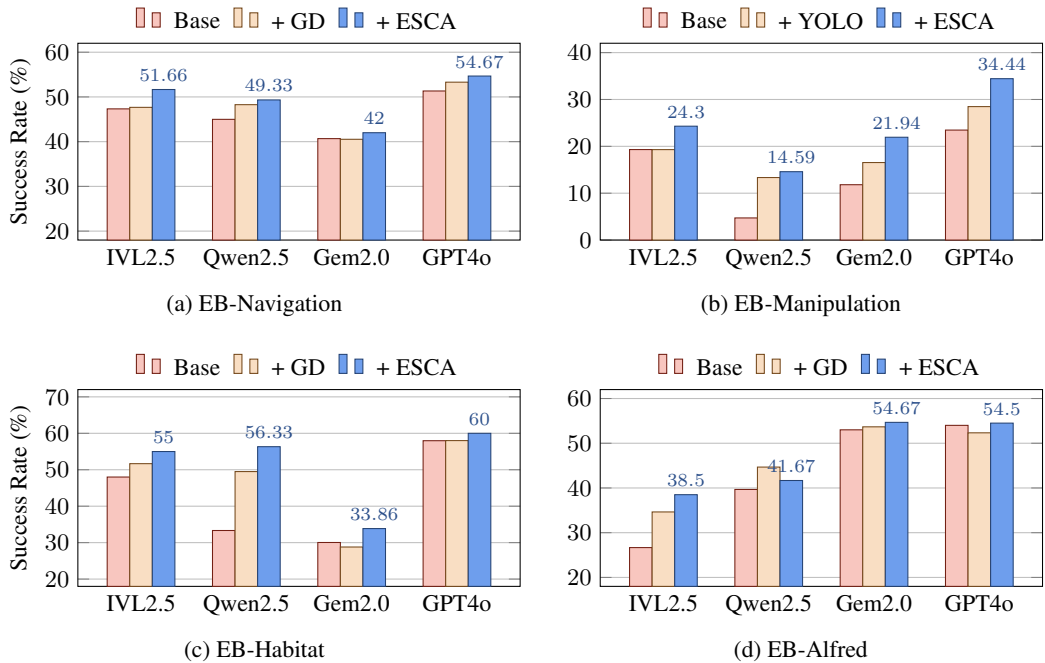

%% file: figures/esca-analysis.tex
\begin{figure}
    \footnotesize
    \begin{minipage}{0.65\linewidth}
        \includegraphics[width=\linewidth]{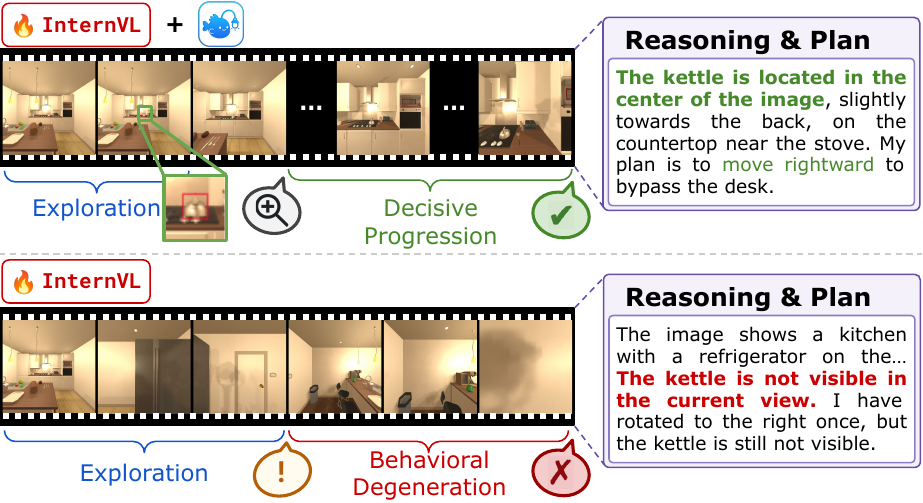}
        \captionof{figure}{
            Two traces by InternVL with and without \ours, on the task of ``navigate to the kettle on the stove''.
            With \ours, the kettle can be marked in the image and textual description, encouraging the agent to decisively move towards the goal.
        }
        \label{fig:qualitative-example}
    \end{minipage}
    \hfill
    \begin{minipage}{0.32\linewidth}
        \centering
        
        
        \newcommand{\donutchart}[3]{
            \def\radiusfirst{0.2cm}
            \def\radiussecond{0.7cm}
            \def\radiusthird{1.2cm}
        
            \pgfmathsetlengthmacro\innerradius{\radiussecond}
            \pgfmathsetlengthmacro\outerradius{\radiusthird}
            \pgfmathsetlengthmacro{\centerradius}{(\outerradius + \innerradius)/2}
            \pgfmathsetlengthmacro{\donutcenter}{\innerradius/2}
            
            \pgfmathsetmacro{\totalnum}{0}
            \foreach \value/\colour/\name/\textcolor in {#2} {
                \pgfmathparse{\value+\totalnum}
                \global\let\totalnum=\pgfmathresult
            }
            
            \pgfmathsetmacro{\wheelwidth}{\outerradius-\innerradius}
            \pgfmathsetmacro{\midradius}{(\outerradius+\innerradius)/2}
            
            \begin{scope}[rotate=90]
                \pgfmathsetmacro{\cumnum}{0}
                \foreach \value/\colour/\name/\textcolor in {#2} {
                    \pgfmathsetmacro{\newcumnum}{\cumnum + \value/\totalnum*360}
                    
                    \pgfmathsetmacro{\midangle}{-(\cumnum+\newcumnum)/2}
                    
                    \filldraw[draw=white,fill=\colour] (-\cumnum:\outerradius) arc (-\cumnum:-(\newcumnum):\outerradius) --
                    (-\newcumnum:\innerradius) arc (-\newcumnum:-(\cumnum):\innerradius) -- cycle;
                    
                    \global\let\cumnum=\newcumnum
                }
            \end{scope}
            
            \pgfmathsetlengthmacro\innerradius{\radiusfirst}
            \pgfmathsetlengthmacro\outerradius{\radiussecond}
            \pgfmathsetlengthmacro{\centerradius}{(\outerradius + \innerradius)/2}
            \pgfmathsetlengthmacro{\donutcenter}{\innerradius/2}
            
            \pgfmathsetmacro{\totalnum}{0}
            \foreach \value/\colour/\name/\textcolor in {#3} {
                \pgfmathparse{\value+\totalnum}
                \global\let\totalnum=\pgfmathresult
            }
            
            \pgfmathsetmacro{\wheelwidth}{\outerradius-\innerradius}
            \pgfmathsetmacro{\midradius}{(\outerradius+\innerradius)/2}
            
            \begin{scope}[rotate=90]
                \pgfmathsetmacro{\cumnum}{0}
                \foreach \value/\colour/\name/\textcolor in {#3} {
                    \pgfmathsetmacro{\newcumnum}{\cumnum + \value/\totalnum*360}
                    
                    \pgfmathsetmacro{\midangle}{-(\cumnum+\newcumnum)/2}
                    
                    \filldraw[draw=white,fill=\colour] (-\cumnum:\outerradius) arc (-\cumnum:-(\newcumnum):\outerradius) --
                    (-\newcumnum:\innerradius) arc (-\newcumnum:-(\cumnum):\innerradius) -- cycle;
                    
                    \global\let\cumnum=\newcumnum
                }
            \end{scope}
            
            \pgfmathsetlengthmacro\innerradius{\radiussecond}
            \pgfmathsetlengthmacro\outerradius{\radiusthird}
            \pgfmathsetlengthmacro{\centerradius}{(\outerradius + \innerradius)/2}
            \pgfmathsetlengthmacro{\donutcenter}{\innerradius/2}
            
            \pgfmathsetmacro{\totalnum}{0}
            \foreach \value/\colour/\name/\textcolor in {#2} {
                \pgfmathparse{\value+\totalnum}
                \global\let\totalnum=\pgfmathresult
            }
            
            \pgfmathsetmacro{\wheelwidth}{\outerradius-\innerradius}
            \pgfmathsetmacro{\midradius}{(\outerradius+\innerradius)/2}
            
            \begin{scope}[rotate=90]
                \pgfmathsetmacro{\cumnum}{0}
                \foreach \value/\colour/\name/\textcolor in {#2} {
                    \pgfmathsetmacro{\newcumnum}{\cumnum + \value/\totalnum*360}
                    
                    \pgfmathsetmacro{\midangle}{-(\cumnum+\newcumnum)/2}
                    
                    \draw[fill=none] node [font=\tiny, color=\textcolor] at (\midangle:{\innerradius+\wheelwidth/2}) {\name};
                    
                    \global\let\cumnum=\newcumnum
                }
            \end{scope}
            
            \pgfmathsetlengthmacro\innerradius{\radiusfirst}
            \pgfmathsetlengthmacro\outerradius{\radiussecond}
            \pgfmathsetlengthmacro{\centerradius}{(\outerradius + \innerradius)/2}
            \pgfmathsetlengthmacro{\donutcenter}{\innerradius/2}
            
            \pgfmathsetmacro{\totalnum}{0}
            \foreach \value/\colour/\name/\textcolor in {#3} {
                \pgfmathparse{\value+\totalnum}
                \global\let\totalnum=\pgfmathresult
            }
            
            \pgfmathsetmacro{\wheelwidth}{\outerradius-\innerradius}
            \pgfmathsetmacro{\midradius}{(\outerradius+\innerradius)/2}
            
            \begin{scope}[rotate=90]
                \pgfmathsetmacro{\cumnum}{0}
                \foreach \value/\colour/\name/\textcolor in {#3} {
                    \pgfmathsetmacro{\newcumnum}{\cumnum + \value/\totalnum*360}
                    
                    \pgfmathsetmacro{\midangle}{-(\cumnum+\newcumnum)/2}
                    
                    \draw[fill=none] node [font=\scriptsize, color=\textcolor, align=center] at (\midangle:{\innerradius+\wheelwidth/2}) {\name};
                    
                    \global\let\cumnum=\newcumnum
                }
            \end{scope}
        }


        \begin{tikzpicture}
            \donutchart{+ ESCA}{
                1/deepseek2!30!white/Ha/deepseek2!60!black, 
                6/deepseek2!30!white/WR/deepseek2!60!black,
                5/llamaOrange!30!white/SR/llamaOrange!60!black, 
                1/llamaOrange!30!white/RE/llamaOrange!60!black, 
                5/gpt1!30!white/IA/gpt1!60!black,
                5/gpt1!30!white/Co/gpt1!60!black%
            }{
                7/deepseek2!60!white/Perc.\\(30\%)/deepseek2!40!black,
                6/llamaOrange!60!white/Reas.\\(26\%)/llamaOrange!40!black,
                10/gpt1!60!white/Plan.\\(44\%)/gpt1!40!black%
            }
            \draw node [align=center] at (-2cm,0) {InternVL\\+ ESCA};
        \end{tikzpicture}


        \begin{tikzpicture}
            \donutchart{Base}{
                2/deepseek2!30!white/Ha/deepseek2!60!black, 
                23/deepseek2!30!white/WR/deepseek2!60!black,
                3/llamaOrange!30!white/SR/llamaOrange!60!black, 
                2/llamaOrange!30!white/RE/llamaOrange!60!black, 
                5/gpt1!30!white/IA/gpt1!60!black,
                2/gpt1!30!white/Co/gpt1!60!black%
            }{
                25/deepseek2!60!white/Perc.\\(69\%)/deepseek2!40!black,
                5/llamaOrange!60!white/Reas.\\(11\%)/llamaOrange!40!black,
                7/gpt1!60!white/Plan. (20\%)/gpt1!40!black%
            }
            \draw node [align=center] at (-2cm,0) {InternVL};
        \end{tikzpicture}
        
        \vspace{-3px}
        
        \captionof{figure}{
            Error decomposition\protect\footnotemark~of InternVL with or without \ours, manually inspected on $60$ EB-Navigation tasks.
        }
        \label{fig:error-decomposition}
    \end{minipage}
\end{figure}
\footnotetext{
    The three top-level error types are \underline{Perc}eption, \underline{Reas}oning, and \underline{Plan}ning. 
    The second-level errors are \underline{Ha}llucination, \underline{W}rong \underline{R}ecognition, \underline{S}patial \underline{U}nderstanding, \underline{S}patial \underline{R}easoning, \underline{R}eflection \underline{E}rror, \underline{I}naccurate \underline{A}ction, and \underline{Co}llision.
    For clarity, the figure uses these acronyms to label the different error types.
}

%% file: figures/adaptation_experiments.tex
\begin{figure}[t!]
    \centering
    \footnotesize
    \begin{minipage}{0.65\textwidth}
        \footnotesize
        
        \pgfplotscreateplotcyclelist{customcolors}{
            {draw=deepseek2!80!black},      
            {draw=gpt1!80!black},    
            {draw=llamaOrange!80!black},    
        }
        
        \begin{subfigure}{0.46\linewidth}
            \centering
            \begin{tikzpicture}
                \begin{axis}[
                    width=1.2\linewidth,
                    height=4.2cm,
                    xtick={2,4,7},
                    xticklabels={1K, 10K, 87K},
                    ylabel={Recall (\%)},
                    ytick={0, 25, 50, 75, 100},
                    ymin=-5, ymax=80,
                    xmin=1.5, xmax=7.5,
                    y label style={at={(axis description cs:0.15,.5)}},
                    x label style={at={(axis description cs:0.5,.05)}},
                    grid=major,
                    legend style={
                        font=\scriptsize,
                        at={(1.13,1)}, 
                        anchor=south,
                        draw=none,       
                        column sep=0px,  
                    },
                    legend columns=3,
                    cycle list name=customcolors,
                    mark size=2pt,
                ]
                
                    \pgfplotstableread[col sep=comma]{data/sgclip_zs_gen.csv}\datatable
    
                    \addplot+[mark=o, line width=1pt] coordinates {(2,16.86) (4,23.68) (7,23.35)};
                    \addlegendentry{OpenPVSG}
                    
                    \addplot+[mark=triangle, line width=1pt,] coordinates {(2,12.91) (4,17.68) (7,17.53)};
                    \addlegendentry{Action Genome}
                    
                    \addplot+[mark=square, line width=1pt,] coordinates {(2,62.16) (4,64.44) (7,71.00)};
                    \addlegendentry{VidVRD}
    
                    \addplot+[dashed, draw=deepseek2!80!black, thick, domain=0:10] {16.33};
                    \node[anchor=south east, font=\scriptsize, color=deepseek2!80!black] at (axis cs:7.5,22.33) {OpenPVSG};
    
                    \addplot+[dashed, draw=gpt1!80!black, thick, domain=0:10] {11.87};
                    \node[anchor=north east, font=\scriptsize, color=gpt1!70!black] at (axis cs:7.5,11.6) {Action Genome};
    
                    \addplot+[dashed, draw=llamaOrange!80!black, thick, domain=0:10] {62.67};
                    \node[anchor=north east, font=\scriptsize, color=llamaOrange!80!black] at (axis cs:7.5,62.85) {VidVRD};
                \end{axis}
            \end{tikzpicture}
            \caption{Entity class prediction R@1}
            \label{fig:sgclip-zs-gen-unary}
        \end{subfigure}
        \hfill
        \begin{subfigure}{0.53\linewidth}
            \centering
            \begin{tikzpicture}
                \begin{axis}[
                    width=1.1\linewidth,
                    height=4.2cm,
                    xtick={2,4,7},
                    xticklabels={1K, 10K, 87K},
                    ytick={0, 25, 50, 75, 100},
                    ymin=-5, ymax=80,
                    xmin=1.5, xmax=7.5,
                    y label style={at={(axis description cs:0.1,.5)}},
                    x label style={at={(axis description cs:0.5,.05)}},
                    grid=major,
                    legend style={
                        font=\scriptsize,
                        at={(1.2,1)}, 
                        anchor=south,
                        draw=none,       
                        column sep=0px,  
                    },
                    legend columns=3,
                    cycle list name=customcolors,
                    mark size=2pt,
                ]
                
                    \pgfplotstableread[col sep=comma]{data/sgclip_zs_gen.csv}\datatable
    
                    \addplot+[mark=o, line width=1pt] coordinates {
                        (2,42.71) 
                        (4,48.04) 
                        (7,53.46)
                    };
                    
                    \addplot+[mark=triangle, line width=1pt] coordinates {
                        (2, 46.56) 
                        (4,72.54) 
                        (7,73.52)
                    };
                    
                    \addplot+[mark=square, line width=1pt] coordinates {
                        (2,27.68) 
                        (4,28.48) 
                        (7,29.41)
                    };
    
                    \addplot+[dashed, draw=deepseek2!80!black, thick, domain=0:10] {9.88};
                    \node[anchor=north east, font=\scriptsize, color=deepseek2!80!black] at (axis cs:7.5,10.88) {OpenPVSG};
    
                    \addplot+[dashed, draw=gpt1!80!black, thick, domain=0:10] {54.78};
                    \node[anchor=south east, font=\scriptsize, color=gpt1!70!black] at (axis cs:7.5,53.6) {Action Genome};
    
                    \addplot+[dashed, draw=llamaOrange!80!black, thick, domain=0:10] {23.45};
                    \node[anchor=north east, font=\scriptsize, color=llamaOrange!80!black] at (axis cs:7.5,24.85) {VidVRD};
                \end{axis}
            \end{tikzpicture}
            \caption{Binary relation prediction R@10}
            \label{fig:sgclip-zs-gen-binary}
        \end{subfigure}
        \captionof{figure}{
            The zero-shot performance of \ourmodel compared to CLIP (shown in dashed lines) on OpenPVSG, Action Genome, and VidVRD datasets.
            We showcase the Recall@1 metrics on entity class prediction, as well as the Recall@10 metrics on binary relation prediction.
            To illustrate data-efficiency, we include the performance of checkpoints of \ourmodel when trained on 1K, 10K, or 87K (full) portion of \ourdataset.
        }
        \label{fig:sgclip-zs-gen}
    \end{minipage}
    \hfill
    \begin{minipage}{0.33\textwidth}
    
        \pgfplotscreateplotcyclelist{customcolors}{
            {draw=gpt1!80!black},    
            {draw=deepseek2!80!black},      
            {draw=llamaOrange!80!black},    
        }
        
        \centering
        \begin{tikzpicture}
            \begin{axis}[
                width=1.1\linewidth,
                height=4.2cm,
                xlabel={$\%$ Data used for fine-tuning},
                xtick={1,2,4},
                xticklabels={$0\%$, $1\%$, $5\%$},
                ylabel={Accuracy (\%)},
                ytick={50, 60, 70, 80, 90, 100},
                ymin=68, ymax=103,
                xmin=0.7, xmax=4.3,
                y label style={at={(axis description cs:0.15,.5)}},
                x label style={at={(axis description cs:0.5,.05)}},
                grid=major,
                legend style={
                    font=\scriptsize,
                    at={(0.5,1)}, 
                    anchor=south,
                    draw=none,       
                    column sep=0px,  
                },
                legend columns=3,
                cycle list name=customcolors,
                mark size=2pt,
            ]
            
                \pgfplotstableread[col sep=comma]{data/sgclip_zs_gen.csv}\datatable

                \addplot+[
                    mark=o, 
                    line width=1pt, 
                    nodes near coords,
                    nodes near coords style={
                        font=\scriptsize,
                        color=gpt1!80!black,
                    }
                ] coordinates {
                    (1,76.34) 
                    (2,80.10) 
                    (4,92.10)
                };
                \addlegendentry{\ourmodel}
                
                \addplot+[mark=triangle, line width=1pt] coordinates {
                    (1,74.37) 
                    (2,78.79) 
                    (4,80.02)
                };
                \addlegendentry{CLIP}
                
                \addplot+[dashed, draw=qwqPink, thick, domain=0:10] {79.8};
                \node[anchor=south east, font=\scriptsize, color=qwqPink] at (axis cs:4.3,80.00) {BIKE};

                \addplot+[dashed, draw=qwqPink, thick, domain=0:10] {77.40};
                \node[anchor=north east, font=\scriptsize, color=qwqPink] at (axis cs:4.3,77.40) {Text4vis};

                \addplot+[solid, draw=llamaOrange!70!black, thick, domain=0:10] {95.90};
                \node[anchor=south east, font=\scriptsize, color=llamaOrange!70!black] at (axis cs:4.3,95.30) {InternVL-6B};
            \end{axis}
        \end{tikzpicture}
        \vspace{-7px}
        \captionof{figure}{
            Down-stream fine-tunability on action recognition, evaluated on ActivityNet dataset.
            We also illustrate zero-shot baselines (BIKE and Text4vis) as well as a fully-supervised baseline (InternVL-6B).
        }
        \label{fig:action-recog-finetunability}
    \end{minipage}%
\end{figure}

%% file: sections/6_related_works.tex
\section{Related Works}
\label{sec:background}


\textbf{Scene Graph in planning.}
Scene graphs~\citep{johnson2015image, xu2017scene, huang2025languagemodelplannerformalizer, huang2021scallop} are symbolic representations that encode the semantic structure of an image or video by identifying objects as nodes and their relationships as edges~\citep{lu2016visual, krishna2017visual}. 
They play a central role in a variety of vision-related tasks, including visual question answering~\citep{lee2019visual, hildebrandt2020scene, qian2022scene}, image captioning~\citep{yang2019auto, zhong2020comprehensive}, and image generation~\citep{herzig2020learning, li2019pastegan}. 
More recently, scene graphs have been increasingly adopted in the domain of robotic planning, for enhanced robustness~\citep{hsu2023s, wang2023programmatically}, or verifiable planning~\citep{jiao2022sequential, rana2023sayplan, dai2024optimalscenegraphplanning}. 
To enable seamless integration with multimodal large language model (MLLM) agents, \ours constructs scene graphs from 2D image inputs and dynamically updates them through embodied interaction with the environment.

\textbf{Embodied Agents.}
Embodied agents~\citep{franklin1997autonomous, zhao2024see} are autonomous systems that perceive, reason, and interact within physical or simulated environments.
To evaluate their capabilities, varies benchmarks~\citep{li2024embodied, xia2018gibson} have been proposed, ranging from vision-language navigation~\citep{savva2019habitat, batra2020objectnav,  deitke2020robothor} and object manipulation~\citep{ehsani2021manipulathor, lin2021softgym} to interactive instruction following and long-horizon planning~\citep{shridhar2020alfred, shridhar2020alfworld, fan2022minedojo}.
Recent advances in large language models (LLMs)~\citep{bommasani2021opportunities, hurst2024gpt, team2023gemini, touvron2023llama} and multimodal large language models (MLLMs)~\citep{bai2023qwen, chen2024internvl, lin2023video, wu2023visual} are driving progress toward general-purpose embodied agents~\citep{ajay2023compositional, belkhale2024rt, brohan2023rt, huang2025limitlanguagemodelsplanning, zhang2023building, ma2024eureka}.
Furthermore, recent MLLMs have been trained end-to-end to directly generate low-level numerical control commands~\citep{brohan2022rt, black2410pi0}.
\ours introduces a general framework for augmenting vision-driven, MLLM-based agents with structured scene graph information.

\textbf{Neurosymbolic Methods with LLM and MLLM.}
A growing trend for enhancing the reasoning capabilities and robustness of large language models (LLMs) is to incorporate structured representations and leverage symbolic algorithms to reason over them~\citep{kassner2021beliefbank, gao2023pal, zhang2023improved, li2024relational, li2025iris, huang2023scallop, relation2024li, solkobreslin2024ised, biberstein2025lobster}. 
These efforts span diverse domains, including code generation~\citep{finnie2022robots, olausson2023linc, liu2024exploring}, mathematical problem solving~\citep{zhang2024mathverse, didolkar2024metacognitive}, and verifiable planning~\citep{silver2022pddl, liu2023llmpempoweringlargelanguage, capitanelli2024framework, zhang2024pddlego}. 
Recent work has extended this paradigm to the low level control domain, training MLLMs with structured scene graphs in an end-to-end manner~\citep{black2410pi0, shi2024composing, qian2024task, zheng2025neurostrata}. 
\ours follows this neurosymbolic direction by introducing \ourmodel, which is trained in MLLM-augmented self-supervised manner enabled by neuro-symbolic methodology, using structured scene graphs as an intermediate representation to guide learning.

%% file: sections/7_conclusion.tex
\section{Conclusion and Limitations}
\label{sec:conclusion}

We introduced \ours, a framework for contextualizing embodied agents through scene graph generation, powered by \ourmodel, a promptable, open-domain scene graph model. 
Through a general transfer protocol, \ours adapts to diverse tasks and consistently improves agent performance across multiple environments and MLLMs. 
Beyond embodied tasks, \ourmodel demonstrates strong generalization and adaptability on open-domain scene graph and action recognition benchmarks.

\textbf{Limitations.} 
Despite its strong performance, our framework has several limitations. 
First, the use of large language models for high-level planning introduces latency, making it unsuitable for real-time low-level control. 
Second, the system relies on 2D visual inputs and lacks support for 3D representations like point clouds, limiting depth-aware reasoning and spatial precision. 
Finally, while \ours leverages MLLMs to generate coherent plans, it lacks formal mechanisms for verifying intermediate and final states during execution.


\textbf{Acknowledgements.} This research was supported by the ARPA-H program on Safe and Explainable AI under award \#D24AC00253-00, the NSF under award \#2313010, Google Research Award, and a gift from AWS AI to ASSET--Penn Engineering Center on Trustworthy AI.

%% file: sections/checklist.tex
\section{Checklist}

\begin{enumerate}

\item {\bf Claims}
    \item[] Question: Do the main claims made in the abstract and introduction accurately reflect the paper's contributions and scope?
    \item[] Answer: \answerYes{} 
    \item[] Justification: We have made four claims in the introduction. All of them are supported by our pipeline and experimental results.

\item {\bf Limitations}
    \item[] Question: Does the paper discuss the limitations of the work performed by the authors?
    \item[] Answer: \answerYes{}.
    \item[] Justification: We have an explicit section dedicated to conclusion, limitation, and future works.

\item {\bf Theory assumptions and proofs}
    \item[] Question: For each theoretical result, does the paper provide the full set of assumptions and a complete (and correct) proof?
    \item[] Answer: \answerNA{}.
    \item[] Justification: We do not include theoretical results.

    \item {\bf Experimental result reproducibility}
    \item[] Question: Does the paper fully disclose all the information needed to reproduce the main experimental results of the paper to the extent that it affects the main claims and/or conclusions of the paper (regardless of whether the code and data are provided or not)?
    \item[] Answer: \answerYes{}.
    \item[] Justification: We include the experimental details in the appendix, and our full code base in supplementary material. 

\item {\bf Open access to data and code}
    \item[] Question: Does the paper provide open access to the data and code, with sufficient instructions to faithfully reproduce the main experimental results, as described in supplemental material?
    \item[] Answer: \answerYes{}.
    \item[] Justification: In the supplementary material, we provide our codebase for (1) preprocessing the \ourdataset dataset and (2) implementing the \ours pipeline and transfer protocols. We will release the dataset and open-source the code upon paper acceptance.

\item {\bf Experimental setting/details}
    \item[] Question: Does the paper specify all the training and test details (e.g., data splits, hyperparameters, how they were chosen, type of optimizer, etc.) necessary to understand the results?
    \item[] Answer: \answerYes{}.
    \item[] Justification: We include these specifications in the appendix.

\item {\bf Experiment statistical significance}
    \item[] Question: Does the paper report error bars suitably and correctly defined or other appropriate information about the statistical significance of the experiments?
    \item[] Answer: \answerNo{}.
    \item[] Justification: We evaluate large language models with the temperature set to 0 to reduce output variance. Moreover, querying these models multiple times with the same prompt is neither environmentally sustainable nor economically feasible.

\item {\bf Experiments compute resources}
    \item[] Question: For each experiment, does the paper provide sufficient information on the computer resources (type of compute workers, memory, time of execution) needed to reproduce the experiments?
    \item[] Answer:  \answerYes{}.
    \item[] Justification: We include the experimental details in the appendix.
    
\item {\bf Code of ethics}
    \item[] Question: Does the research conducted in the paper conform, in every respect, with the NeurIPS Code of Ethics \url{https://neurips.cc/public/EthicsGuidelines}?
    \item[] Answer: \answerYes{}.
    \item[] Justification: No human objectives are involved in our research. Our dataset is created on the top of a fully open source dataset with Apache License 2.0. 

\item {\bf Broader impacts}
    \item[] Question: Does the paper discuss both potential positive societal impacts and negative societal impacts of the work performed?
    \item[] Answer: \answerNA{}.
    \item[] Justification:  This work focuses on evaluating embodied agent performance in simulated environments, with accuracy measured against predefined, human-authored criteria. As such, the study does not directly engage with real-world deployment or societal impact, making a discussion of positive or negative societal effects not applicable in this context.
    
\item {\bf Safeguards}
    \item[] Question: Does the paper describe safeguards that have been put in place for responsible release of data or models that have a high risk for misuse (e.g., pretrained language models, image generators, or scraped datasets)?
    \item[] Answer: \answerNA{}.
    \item[] Justification: Our paper poses no such risks.

\item {\bf Licenses for existing assets}
    \item[] Question: Are the creators or original owners of assets (e.g., code, data, models), used in the paper, properly credited and are the license and terms of use explicitly mentioned and properly respected?
    \item[] Answer: \answerYes{}.
    \item[] Justification: The LLava-Video-178K dataset is under Apache 2.0 license.

\item {\bf New assets}
    \item[] Question: Are new assets introduced in the paper well documented and is the documentation provided alongside the assets?
    \item[] Answer: \answerYes{}.
    \item[] Justification: we provide the whole repository that generates our dataset. 

\item {\bf Crowdsourcing and research with human subjects}
    \item[] Question: For crowdsourcing experiments and research with human subjects, does the paper include the full text of instructions given to participants and screenshots, if applicable, as well as details about compensation (if any)? 
    \item[] Answer: \answerNA{}.
    \item[] Justification: the paper does not involve crowdsourcing nor research with human subjects.

\item {\bf Institutional review board (IRB) approvals or equivalent for research with human subjects}
    \item[] Question: Does the paper describe potential risks incurred by study participants, whether such risks were disclosed to the subjects, and whether Institutional Review Board (IRB) approvals (or an equivalent approval/review based on the requirements of your country or institution) were obtained?
    \item[] Answer: \answerNA{}.
    \item[] Justification: the paper does not involve crowdsourcing nor research with human subjects.

\item {\bf Declaration of LLM usage}
    \item[] Question: Does the paper describe the usage of LLMs if it is an important, original, or non-standard component of the core methods in this research? Note that if the LLM is used only for writing, editing, or formatting purposes and does not impact the core methodology, scientific rigorousness, or originality of the research, declaration is not required.
    \item[] Answer: \answerYes{}.
    \item[] Justification: \ours is a new method augmenting MLLMs with scene graphs.

\end{enumerate}

%% file: sections/appendix.tex
\newpage
\appendix

\input{sections/app_1_dataset}
\newpage

\input{sections/app_2_training}
\newpage

\input{sections/app_3_experimental}

\newpage

%% file: sections/app_1_dataset.tex
\section{\ourdataset}

\subsection{Video and Caption Source}
We build upon the LLaVA-Video-178K dataset~\cite{zhang2024videoinstructiontuningsynthetic}, which comprises diverse video sources and GPT-generated, detailed video descriptions. 

\ourdataset contains a total of $87,045$ datapoints, each consisting of a video clip ranging from $0$ to $30$ seconds in length, along with its corresponding caption. 
The videos are drawn from ten primary sources, spanning a wide range of content—including egocentric and household activities, as well as crowd-sourced videos from YouTube. 
Specifically, these ten sources are HD-VILA-100M~\cite{xue2022advancing}, InternVid-10M~\cite{wang2023internvid}, VidOR~\cite{shang2019annotating}, VIDAL (YouTube Shorts)~\cite{zhu2024vidal}, YouCook2~\cite{zhou2017youcook2}, Charades~\cite{sigurdsson2016charades}, ActivityNet~\cite{caba2015activitynet}, Kinetics-700~\cite{kay2017kinetics}, Something-Something v2~\cite{goyal2017something}, and Ego4d~\cite{grauman2022ego4d}. 
The captions were generated using GPT-4 with a dense frame sampling technique \cite{zhang2024videoinstructiontuningsynthetic}.

\input{figures/app_mask_gen_algorithm}

The two main contributions of \ourdataset{} are the inclusion of object trajectories and executable programmatic specifications. 
We leverage SAM2~\cite{ravi2024sam2segmentimages} to generate object trajectories and design a custom algorithm to handle objects that do not appear in the first frame. 
To obtain the specifications, we use GPT-4 and develop a transcompiler that converts them into executable programs.

These two additional components enable fine-grained spatio-temporal alignment between the video content and the concepts expressed in the specifications~\cite{huang2025laser}. Notably, the only source of supervision is the video itself; all other annotations are derived using multimodal large language models (MLLMs). We refer to this approach as \emph{model-driven self-supervision}.

\newpage
\subsection{Object Trajectory Generation}

We aim to generate object trajectories from videos using Segment Anything 2 (SAM2).
A key challenge is discovering new objects that do not appear in the first frame, which is not natively supported by SAM2.
To address this, we design an iterative algorithm that identifies the next frame most likely to contain a new object and propagates its mask throughout the video, as illustrated in \algref{alg:prompt-mask-propagation}. 
To further leverage concepts generated by GPT-4, we extend this algorithm to incorporate arbitrary frame-based bounding box generators, such as Grounding DINO~\cite{liu2024groundingdinomarryingdino} and YOLO~\cite{hidayatullah2025yolo11}.
This extended version uses a prompt scheduler that prioritizes frames containing the highest number of grounded objects and iteratively propagates them across the video, as shown in \algref{alg:gdc-prompt-buffer} and \algref{alg:generate-masks-grounding-dino}.

\newpage
\subsection{Concept Generation}

To construct low-level supervision from high-level captions, we leverage GPT-4 to generate spatio-temporal specifications.
To mitigate potential hallucinations from the language model, we design a compiler that verifies the validity of the generated programs.
We use a few-shot prompt with three examples. 

\textbf{Prompt 1: Role Definition}
\begin{lstlisting}[language=json, numbers=none]
You are a super user in logic programming. 
You are also an expert at structured data extraction.
\end{lstlisting}

\textbf{Prompt 2: General Task Instruction}
\begin{lstlisting}[language=json, numbers=none]
You will describe the event length and location in 
both natural language and fraction of the video. 
The natural language description of the locations in the video can be: early, mid, late.
The natural language description of the durations of the event can be: long, medium, short
Examples of precise video locations: [1/4, 1/2], [2/3, 1].
Examples of event durations: 1/4, 2/3, 1.
\end{lstlisting}

\textbf{Prompt 3: Few-shot Examples}

\begin{minipage}{\textwidth}
\centering
\begin{lstlisting}[language=json, numbers=none, caption="One of the few-shot example for concept extraction"]
Caption: A man carries a child and walks to the left from behind a woman holding another child.
Video_ID: "0be30efe",
Action json: 
{"video_id": "0be30efe", 
 "sequential descriptions": [ "man A carry child B, women C hold child D, man A is behind women C", "man A walk", "man A at left" ], 
  "time stamps": 
    { "1": { "description": [ "man A carry child B", "women C hold child D", "man A is behind women C" ], "programmatic": [ "carrying(A, B)", "name(A, man)", "name(B, child)", "holding(C, D)", "name(C, women)", "name(D, man)", "behind(A, C)",  ], "duration": "short", "duration precise": "1/4", "video location": "early",  "video location precise": "[0, 1/4]" }, "2": { "description": [ "man A walk" ], "programmatic": [ "walk(A)", ], "duration": "medium", "duration precise": "1/2", "video location": "mid", "video location precise": "[1/4, 3/4]" }, "3": { "description": [ "man A at left" ], "programmatic": [ "left(A)" ], "duration": "short", "duration precise": "1/4", "video location": "late", "video location precise": "[3/4, 1]"}}}
\end{lstlisting}
\end{minipage}

\textbf{Prompt 4: Concept Definition}
\begin{lstlisting}[language=json, numbers=none]
Note all the relations are name, unary or binary.
A name relation takes in two arguments, the first is always a variable, and the second argument could be noun ("apple"), noun phrase ("ancient_building"), location("dark_forest"), etc. For example "name(A, "apple")" means the variable A refers to an apple. Please ensure no space occur in the second argument.
A unary relation takes in one variable as its argument. For example, close(A) means A is close to the camera.
A binary relation takes in two variables as its arguments. For example, above(A, B) means A is above B.
The entity in the binary and unary relation are variables in the form of capitalized letters (A, B).
The predicate of the unary relation can be adjectives, verbs, and name.
The predicate of the binary relation can be preposition, and verb. 
Please include the name relation any time if applicable.
Please make the preposition and adjectives into separate two relation. 
For each time stamp, please only describe the events that are happening at the same time, if any sequential events occur, put them into multiple time stamps. 
For example, instead of 'person A enter from left, person A walk to center, person A move to couch' in one time stamp, 
put it into three different time stamps: "person A enter from left", "person A walk to center", "person A move to couch".
Please only describe one single event in sequential description per time stamp. 
Please use as many relations as possible to precisely describe the action.
Please generate the action json programs for the following captions in the following format:
{"actions": {caption_id: action json programs}}
IMPORTANT: Please REMOVE all new line characters and extra spaces in the generated json!
\end{lstlisting}

\newpage
\subsection{Dataset Statistics}
\begin{figure}[h!]
    \centering

    \begin{subfigure}[t]{0.45\textwidth}
        \centering
        \includegraphics[width=\linewidth]{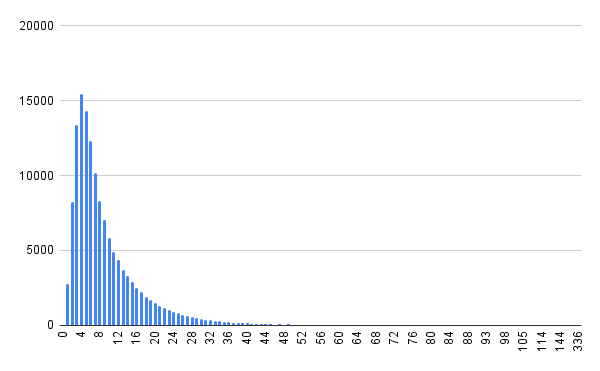}
        \caption{Name Count}
    \end{subfigure}
    \hfill
    \begin{subfigure}[t]{0.45\textwidth}
        \centering
        \includegraphics[width=\linewidth]{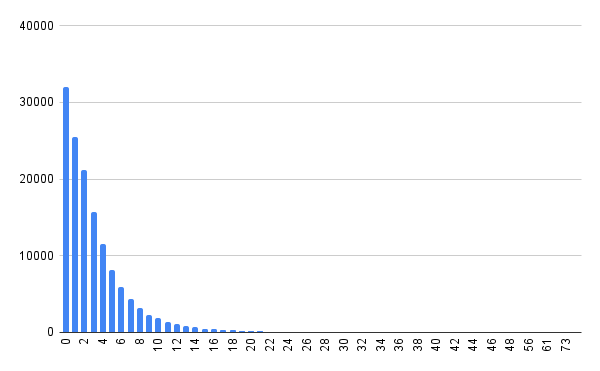}
        \caption{Action Count}
    \end{subfigure}

    \vspace{1em} 

    \begin{subfigure}[t]{0.45\textwidth}
        \centering
        \includegraphics[width=\linewidth]{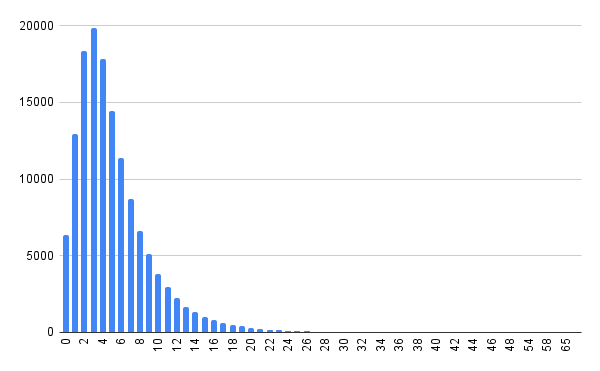}
        \caption{Relation Count}
    \end{subfigure}
    \hfill
    \begin{subfigure}[t]{0.45\textwidth}
        \centering
        \includegraphics[width=\linewidth]{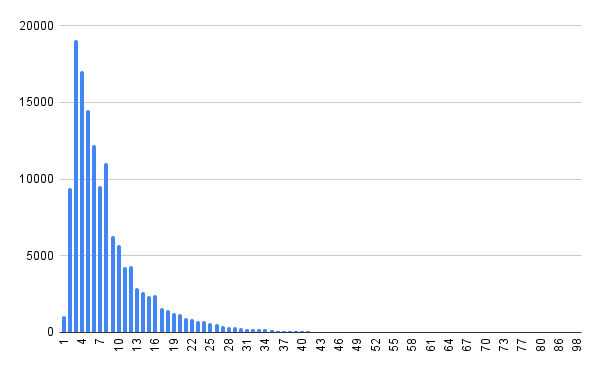}
        \caption{Event Count}
    \end{subfigure}

    \caption{Distribution of Names, Attributes, Relations, and Events}
    \label{fig:count-distributions}
\end{figure}

We present the statistics of \ourdataset through three perspectives: the composition of video sources, the complexity of extracted specifications, and the word clouds of associated concepts.
We selected 0–30 second clips from the \textsc{LLaVA-Video-178K} dataset~\cite{zhang2025llavavideovideoinstructiontuning}, with the resulting source distribution shown in \figref{fig:stats_video_source}.
Our extracted spatio-temporal specifications exhibit considerable complexity.
As illustrated in \figref{fig:count-distributions}, a single specification can contain multiple names, actions, relations, and events.
In addition, we visualize the vocabulary diversity using word clouds for names, actions, and relations.
In total, our dataset includes 220{,}905 unique names, 57{,}930 unique actions, and 35{,}415 unique relation keywords.

\begin{figure}
    \centering
    \begin{tikzpicture}
    \tikzset{
      lines/.style={draw=white, line width=0.6pt},
      pie/label/.style={font=\small, text=black},
    }
    \pie[
  color={
    blue!65!white,
    teal!60!white,
    orange!75!white,
    purple!60!white,
    red!55!white,
    gray!65!white
  },
  sum=auto,
  after number=\%,
  text=pin,
  every pin/.style={pie/label, pin distance=1.2em},
  every only number node/.style={pie/label},
  style={lines},
  rotate=30,
  explode=0.03
]{
  86.99/Youtube,
  4.05/Charades,
  1.47/NextQA,
  6.40/Youcook2,
  0.96/ActivityNet,
  0.14/Ego4D
}
\end{tikzpicture}
    \caption{Video sources for \ourdataset.}
    \label{fig:stats_video_source}
\end{figure}

\begin{figure}[h!]
    \centering
    \begin{subfigure}[t]{0.32\textwidth}
        \centering
        \includegraphics[width=\linewidth]{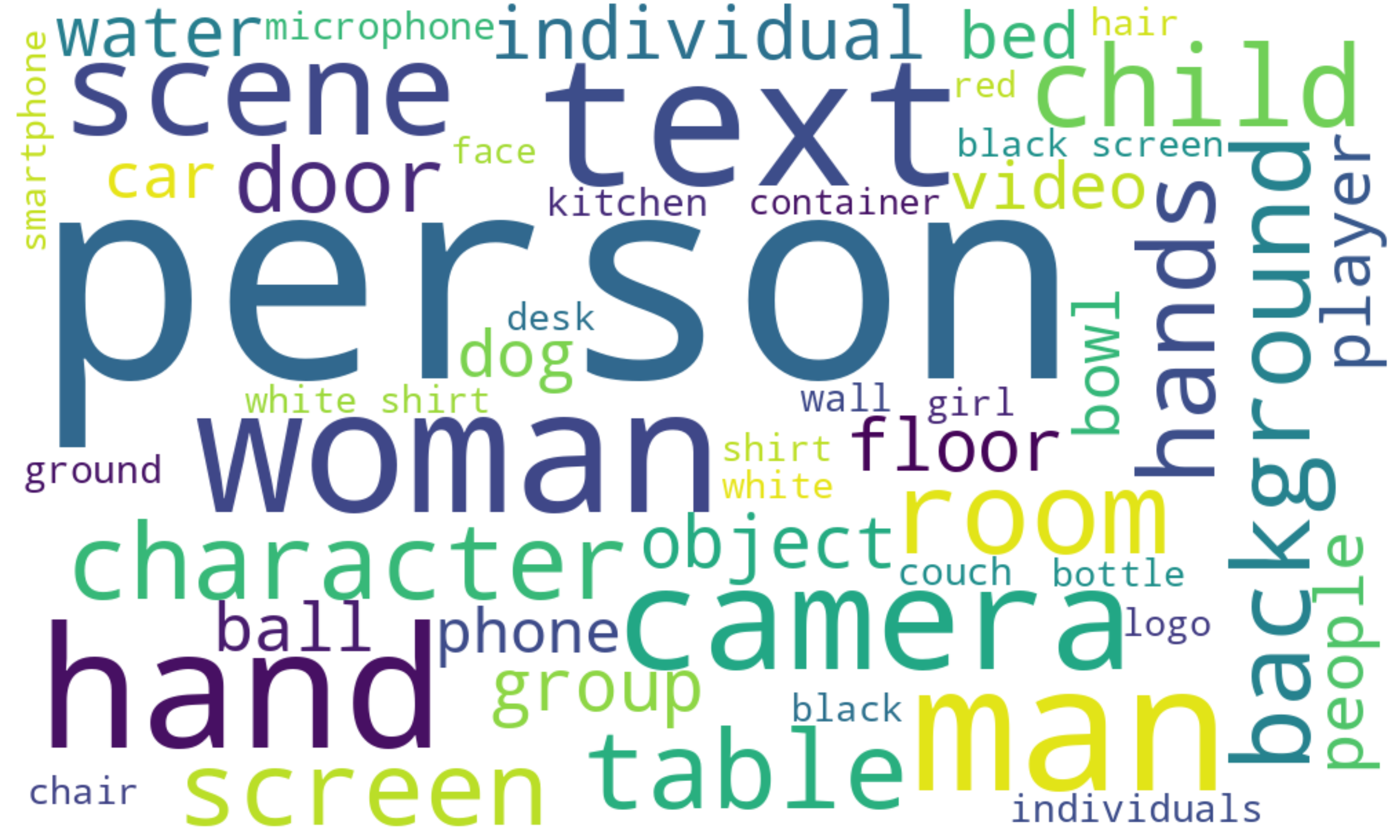}
        \caption{Name}
    \end{subfigure}
    \hfill
    \begin{subfigure}[t]{0.32\textwidth}
        \centering
        \includegraphics[width=\linewidth]{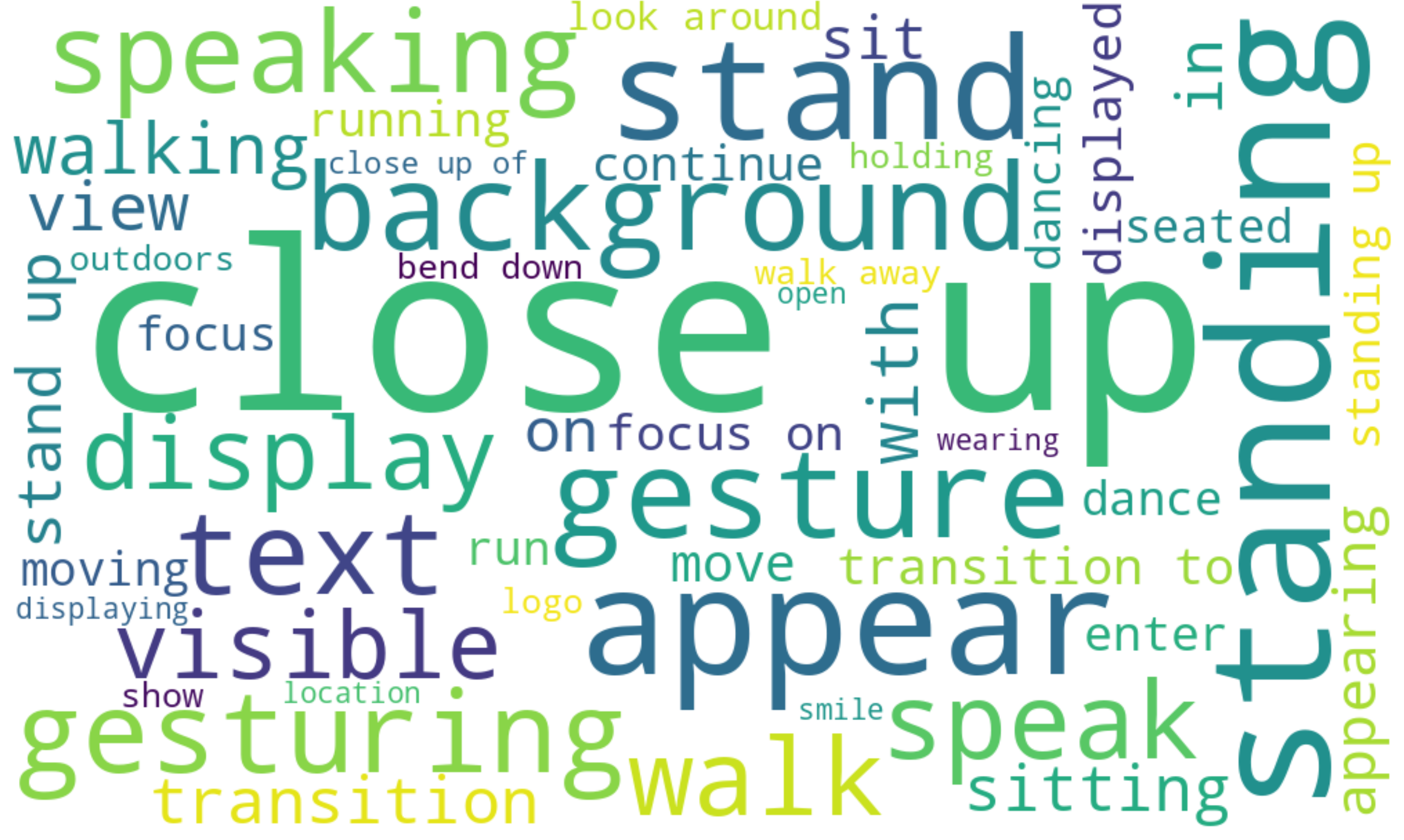}
        \caption{Action}
    \end{subfigure}
    \hfill
    \begin{subfigure}[t]{0.32\textwidth}
        \centering
        \includegraphics[width=\linewidth]{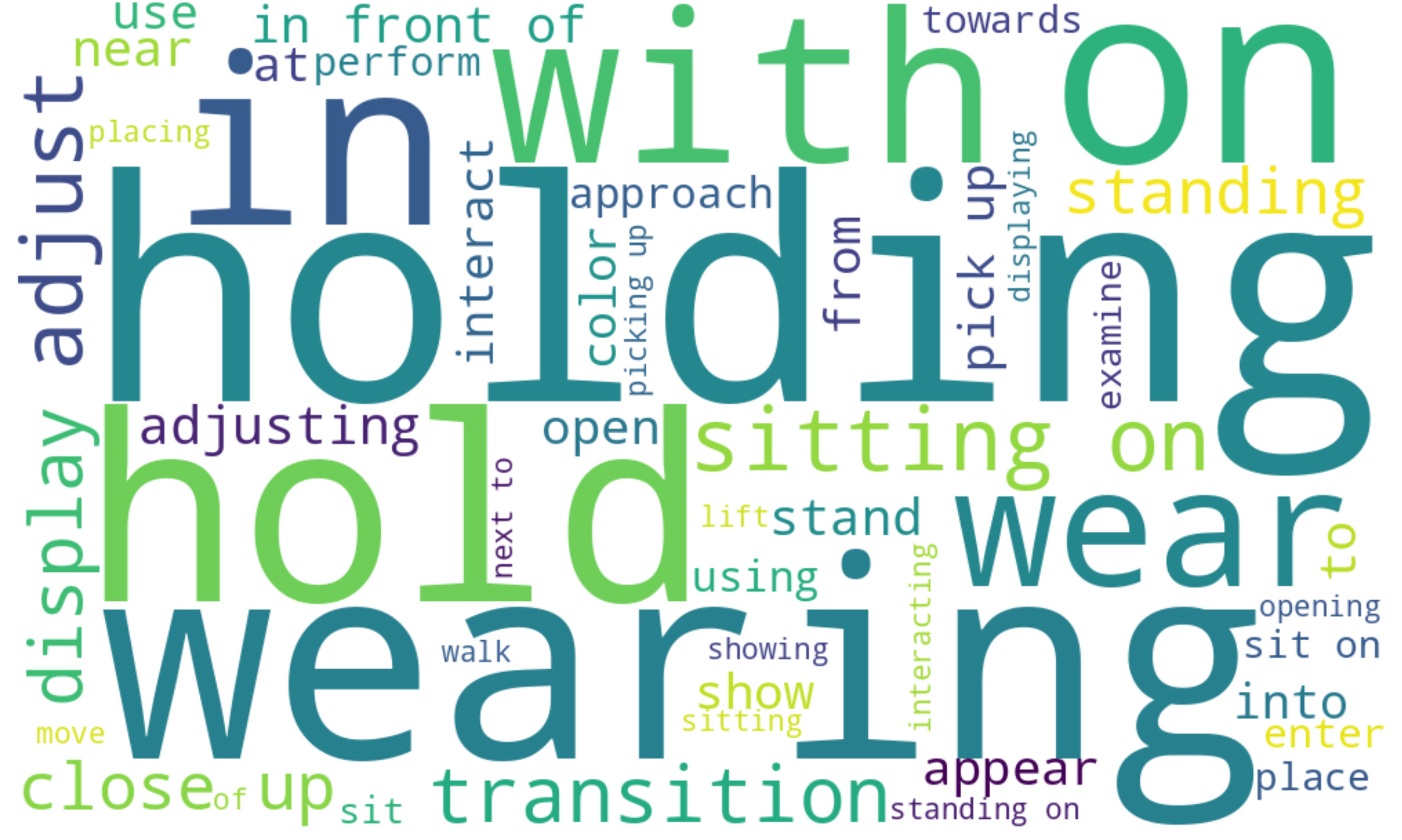}
        \caption{Relation}
    \end{subfigure}
    \caption{Word cloud visualizations of the top-$50$ most frequent names, attributes, and relations.}
    \label{fig:wordclouds}
\end{figure}

%% file: figures/app_mask_gen_algorithm.tex
\begin{algorithm}[t]
\caption{Video Mask Propagation with New Object Discovery}
\label{alg:prompt-mask-propagation}
\KwIn{Video $V = \{I_0, \dots, I_T\}$, mask generator $\mathcal{G}$, propagation model $\mathcal{P}$, state $S$}
\KwIn{Batch size $B$, thresholds: $\tau_{\text{iou}}, \tau_{\text{score}}, \tau_{\text{inner}}$, stride $s$}
\KwOut{Mask dictionary $\texttt{video\_segments} : \texttt{Dict[Frame][ObjectID] = Mask}$}

\texttt{video\_segments} $\gets \emptyset$, \texttt{now\_frame} $\gets 0$\;

\While{not \texttt{saturated}}{
    $I \gets V[\texttt{now\_frame}]$; \tcp{Load current image frame}
    $\mathcal{M} \gets \mathcal{G}(I)$; \tcp{Generate masks for current frame}
    \tcp{Find new object masks not yet propagated from this frame}
    $\mathcal{M}_{new}$ = filter($\mathcal{M}$,
    \texttt{video\_segements[now\_frame]}, $\tau_{\text{iou}}, \tau_{\text{score}}, \tau_{\text{inner}})$\; 

\tcp{Register the new masks as prompts} $\mathcal{P}_{\text{buf}}$.update($\mathcal{M}_{\text{new}}$)\;

\tcp{Prompt the mask generator with updated masks as initialize condition}
    $\mathcal{P}.\texttt{reset\_state}(S)$\;

    \ForEach{(oid, frame\_id, mask) in $\mathcal{P}_{\text{buf}}$}{
    $\mathcal{P}.\texttt{add\_new\_mask\_prompt}(S, frame\_id, o, m)$\;}
    
\tcp{Propagate the mask prompts through the whole video}         
    \ForEach{$(f, \{o_i\}, \{\ell_i\}) \in \mathcal{P}.\texttt{propagate}(S)$}{
        \If{$f \notin \texttt{video\_segments}$}{
            $\texttt{video\_segments}[f] \gets \{\}$\;
        }
        \ForEach{$i$}{
            $m_i \gets \texttt{binarize}(\ell_i)$\;
            $\texttt{video\_segments}[I_f][o_i] \gets m_i$\;
        }
    }

    \tcp{Find the next frame to process with least mask coverage}
    Compute coverage $\rho_f$ over $f \in \{ \texttt{now\_frame} + 1, \dots T \}$\;
    \texttt{saturated}, \texttt{now\_frame} = check\_saturation($\rho$)
}

\Return $\texttt{video\_segments}$
\end{algorithm}

\begin{algorithm}[t]
\caption{Bounding Box Prompt Buffer Class}
\label{alg:gdc-prompt-buffer}
\KwOut{Prompt dictionary for propagation; object-to-class mapping}

Initialize fields: 
\texttt{oid},
\texttt{frame2bboxes}, 
\texttt{frame2masks}, 
\texttt{valid\_prompts}, 
\texttt{overlapped\_prompts}\;

\vspace{1mm}
\tcp{Add predicted bounding box and mask}
\SetKwFunction{AddPrompt}{AddPrompt}
\Fn{\AddPrompt{$frame\_id$, $box$, $mask$,}}{
    \If{not valid(m)}{\Return}
\texttt{frame2bboxes[$frame\_id$]}.append($box$);
\texttt{frame2masks[$frame\_id$]}.append($mask$)
}

\vspace{1mm}
\SetKwFunction{PopDenseFrame}{PopMostDensePromptFrame}
\Fn{\PopDenseFrame{}}{
    \tcp{Find frame $f^*$ with largest mask area}
    $f^* \gets \arg\max_f \sum \texttt{frame2masks}[f]$\;
    
    \tcp{Assign object IDs starting from \texttt{oid} for each bbox in $f^*$}
    \ForEach{box in frame2bboxes[$f^*$]}{
        \texttt{valid\_prompts[$f^*$][oid]} $\gets$ box;
        oid += 1\;
    }
    \tcp{Remove prompts from to process list}
    \texttt{frame2bboxes}.pop($f^*$); \texttt{frame2masks}.pop($f^*$)\; 
    \Return $f^*, \texttt{valid\_prompts}[f^*]$
}

\vspace{1mm}
\SetKwFunction{RemoveDup}{RemoveDuplicatePrompts}
\Fn{\RemoveDup{$\texttt{video\_segments}$}}{
    \ForEach{ $frame\_id$ in \texttt{video\_segments}}{
        \If{$frame\_id$ not in \texttt{frame2masks}}{\Continue}
        Compute IoU between predicted and prompt masks in $frame\_id$\;
        \tcp{Filter prompts already covered}
        Remove overlapping segments from \texttt{frame2bboxes} and \texttt{frame2masks}
    }
}

\end{algorithm}

\begin{algorithm}[t]
\caption{Mask Generation via Frame-wise Bounding Box Grounding}
\label{alg:generate-masks-grounding-dino}
\KwIn{Grounding model $\mathcal{G}$, SAM predictor $\mathcal{P}$, SAM mask generator $\mathcal{M}$, video frames $\{I_0, \dots, I_T\}$, label set $\mathcal{C}$}
\KwIn{Box and text thresholds $\tau_b, \tau_t$, target FPS, max propagation steps $k$}
\KwOut{Per-frame mask segments \texttt{video\_segments}, object-to-class mapping \texttt{oid\_pred}}

Initialize \texttt{prompt\_memory} $\gets$ \texttt{PromptBuffer()}\;
\tcp{Step 1: Grounding-based object detection}
\ForEach{frame $I_f$ in video}{
    $\texttt{bboxes}$ = $\mathcal{G}$($I_f$, $\mathcal{C}$)\;
    $\texttt{video\_boxes}[f]$ $\gets$ $\texttt{bboxes}$ \;
}

\tcp{Step 2: Generate masks for all bounding boxes}

\ForEach{$(f, \texttt{bboxes}) \in \texttt{video\_boxes}$}{
    masks = $\mathcal{M}$($I_f$, \texttt{bboxes}) \;
    \ForEach{(box, mask) in zip(\texttt{bboxes}, masks)}
    {\texttt{prompt\_memory.add\_prompt}($f, box, mask$)}
}

\tcp{Step 3: Iterative propagation from dense prompt frame}
$f_{\text{start}}, \texttt{prompt} \gets \texttt{prompt\_memory.pop\_most\_dense\_fid\_prompt}()$\;
\texttt{video\_segments} $\gets \emptyset$, $t \gets 0$\;

\While{$\texttt{prompt} \neq \emptyset$ and $f_{\text{start}} \neq -1$ and $t < k$}{
    Reset predictor state $S \gets S_0$\;
    \ForEach{$(f, \texttt{frame\_prompt}) \in \texttt{prompt\_memory.valid\_prompts}$}{
        \ForEach{$(o, b) \in \texttt{frame\_prompt}$}{
            Add prompt $(o, b)$ at frame $f$ into $\mathcal{P}$ with state $S$\;
        }
    }
    
    \tcp{Propagate forward}
    \ForEach{$(f, \{o_i\}, \{\ell_i\}) \in \mathcal{P}.\texttt{propagate}(S)$}{
        \texttt{video\_segments}[$f$] $\gets$ binarized masks $\{\ell_i > 0\}$ mapped to $o_i$
    }

    \tcp{Propagate backward}
    \ForEach{$(f, \{o_i\}, \{\ell_i\}) \in \mathcal{P}.\texttt{propagate}(S, reverse=True)$}{
        \texttt{video\_segments}[$f$] $\gets$ binarized masks $\{\ell_i > 0\}$ mapped to $o_i$
    }

    \texttt{prompt\_memory.remove\_dup\_prompt}(\texttt{video\_segments})\;
    $f_{\text{start}}, \texttt{prompt} \gets \texttt{prompt\_memory.pop\_most\_dense\_fid\_prompt}()$\;
    $t \gets t + 1$\;
}

\Return $\texttt{video\_segments}$
\end{algorithm}

%% file: sections/app_2_training.tex
\section{\ourmodel Training Details}

We build our training pipeline following the LASER work~\cite{huang2025laser}, which uses a weakly-supervised approach to align video with its caption. 
Note that the caption used in \ourmodel is also generated from the video with multi-modal large language model (GPT-4o), and this leads to all labels are generated from the video itself, thus, we consider this approach as a model-driven self-supervised method. 
In this section, we elaborate on the details of how \ourmodel is trained and providing details in the training procedure.

\subsection{Compute}
All our experiments are carried out on a device with
(1) $128$ $32$-core Intel(R) Xeon(R) Gold 6338 CPU @ 2.00GHz 
(2) $10$ NVIDIA H100 PCIe GPUs.
\ourmodel takes $10$ days to finetune on the $87K$ datapoints for $3$ epochs.

\subsection{Loss}
We train the \ourmodel with three different losses: contrastive loss, temporal loss, and semantic loss. 

\textbf{Contrastive Loss.}
To help \ourmodel distinguish between concepts that appear in a video and those that do not, we adopt a contrastive loss design.
For each video $V$, we pair it with its corresponding specification $\phi$ and randomly sample an unrelated specification $\phi'$ from the dataset.
The objective is to maximize the alignment score between the matching pair $(V, \phi)$ toward 1, while minimizing the score between the mismatched pair $(V, \phi')$ toward 0.

A key challenge arises from the fact that specifications are exceptionally long—each containing an average of 8.24 events per datapoint—making it computationally expensive to ground the entire specification in the video.
To address this, we introduce a chunked event training strategy.
Each specification is divided into smaller chunks containing at most $3$ events, and we align each chunk with the full video.
When sampling mismatched pairs, we similarly draw from these smaller chunks to maintain alignment granularity and training efficiency.

\textbf{Temporal Loss.}
To improve alignment accuracy between each event and its corresponding location in the video, we query GPT for a fine-grained estimation of the event's temporal span.
For all satisfiable event sequences, we assign higher rewards to those aligned with the desired temporal location, and reduced rewards to those outside the target range.

\textbf{Semantic Loss.}
We further incorporate common-sense negations to improve the model's understanding of what is unlikely to occur in a given scenario.
For example, if a scene is outdoors, it is unlikely to contain a bed; if a person is “standing,” they are unlikely to be “sleeping” at the same time; if a person is “holding” an object, it is improbable that the object is “hitting” the person simultaneously.

To introduce this notion of negation, we first identify the top $5,000$ most frequent keywords across names, actions, and relations.
Using word vectors from SpaCy~\cite{honnibal2020spacy}, we compute the 50 most semantically distant keywords within this subset for each current scenario.
During training, we sample $5$ keywords per category (name, action, relation) from these $50$ distant terms to compute a semantic loss that penalizes implausible pairings.

\subsection{Hyperparameters}

We use a learning rate of $1 \times 10 ^ {-6}$ and a batch size of 2. The video is sampled at a target frame rate of 1 FPS. For the semantic loss, we sample 5 negative keywords per instance and set the semantic loss weight to 0.1. In the provenance setting for Scallop, we use difftopkproofs with a top-$k$ value of $3$ for proof extraction.
We fine tune from the CLIP model with a total of $3$ epochs, and we evaluate and ensemble on the best performing models from different epochs. 

%% file: sections/app_3_experimental.tex
\section{Additional Experimental Results}

\subsection{EmbodiedBench: Navigation}
\input{figures/performance-eb-nav}

We demonstrate that \ours enhances agent performance in navigating to target objects on embodied benchmarks, as shown in \tabref{tab:embodiedbench_nav_detailed}.
Across four base multimodal large language models, \ours consistently outperforms both the base models and variants using only Grounding-DINO.
We outline the design of the transfer protocol and provide qualitative studies in the following section.

\textbf{Transfer Protocol: Grounding Dino}
The overall pipeline of \ours with Grounding DINO consists of three main steps: concept extraction, object identification, and visual summarization and validation.
Rather than performing full scene graph prediction, we directly use Grounding DINO to extract candidate target objects based on their name and attributes.
Each concept input to Grounding DINO follows the format \texttt{<attribute> <object name>}, such as "a grey rectangular object."
The \ours pipeline then passes the augmented image and the newly constructed query back to the MLLM to predict subsequent actions.
Note that, based on our study, Grounding DINO tends to generate false positive candidate objects, which leads to a performance drop on the long-horizon subset.
This is the only task subset where the target object is not visible in the initial scenario.
The box threshold for grounding DINO is $0.2$, and the text threshold is $0.1$.

\textbf{Transfer Protocol: \ours}
The overall pipeline of \ours with \ourmodel consists of four main steps: concept extraction, object identification, scene graph generation, and visual summarization and validation.
We begin by using Grounding DINO to extract candidate target objects and their related objects based on object names.
Next, \ourmodel predicts the attributes and relationships of the identified objects and aggregates predictions across names, attributes, and relationships.
To ensure precision, we select only the top-$1$ object identified as the target and apply a confidence threshold of $0.3$.
For Grounding DINO, we use a box threshold of $0.1$ and a text threshold of $0.1$.

\textbf{Qualitative Studies:}
We present two tasks to qualitatively analyze performance on the EB-Navigation environment.
Each task is solved using three configurations: GPT-4o alone, GPT-4o augmented with Grounding DINO, and GPT-4o augmented with \ours.
Compared to GPT-4o alone, both Grounding DINO and \ours provide additional information that aids task completion.
However, \ours results in significantly fewer false negatives than Grounding DINO, leading to improved performance on long-horizon tasks.

\begin{figure}
    \centering
    \includegraphics[width=\linewidth]{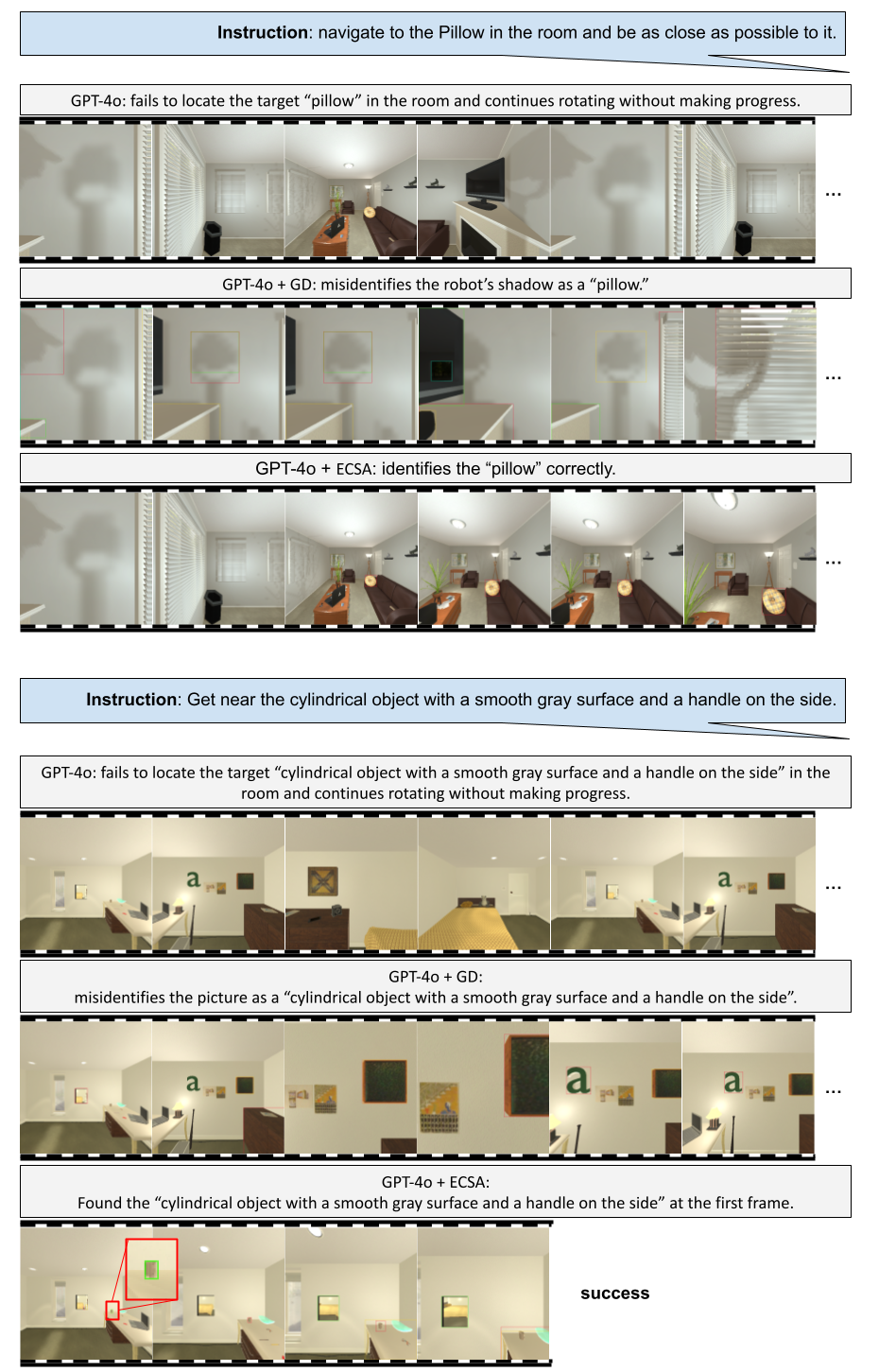}
    \caption{We present two qualitative examples comparing the original MLLM (GPT-4o), its variant augmented with Grounding DINO, and the full \ours pipeline.
Our observations show that \ours improves the precision and quality of target object recognition. }
    \label{fig:ebnav_qualitative}
\end{figure}

\begin{figure}
    \centering
    \input{figures/eb-manip-algorithm}
    \caption{Entropy-guided assignment algorithm that maps entity names to objects based on assignment scores and availability. At each step, it selects the object with the lowest entropy over valid assignments, then assigns the highest scoring available entity. This process continues until all objects are assigned.
    }
    \label{fig:entropy_alg}
\end{figure}

\newpage
\subsection{EmbodiedBench: Manipulation}
\input{figures/performance-eb-mani}

EB-Manipulation evaluates an agent’s ability to perform fine-grained object manipulation using visual input and language instructions. Unlike typical embodied AI settings, it provides full 3D spatial information (X, Y, Z coordinates) for all objects, removing the need for spatial inference and shifting the focus to semantic grounding and planning. 

To construct structured object representations, we use the provided coordinates to generate point-based prompts for SAM 2.1, which produces segmentation masks that are converted into bounding boxes. 
Each object is annotated with a unique integer label starting from 
1. we then prompt a vision-language model (VLM) to generate textual descriptions in the form <color> <object> (e.g., ``yellow star''), using handcrafted templates to ensure consistency and match the number of detected objects. 

Since the VLM outputs are unordered, we use ESCA, which treats predicted entities as categorical labels and outputs probability distributions over them for each bounding box. 
To resolve the correspondence between entities and boxes, we apply an entropy-based matching algorithm (Figure 3), which iteratively assigns the most confident (lowest-entropy) pairings while enforcing uniqueness constraints. 
The resulting assignments are used to generate a structured natural language description (e.g., ``Object 1 is a yellow star''), which is appended to the original prompt and passed to the VLM to support precise and context-aware manipulation planning.

\begin{figure}
    \centering
    \includegraphics[height=20cm]{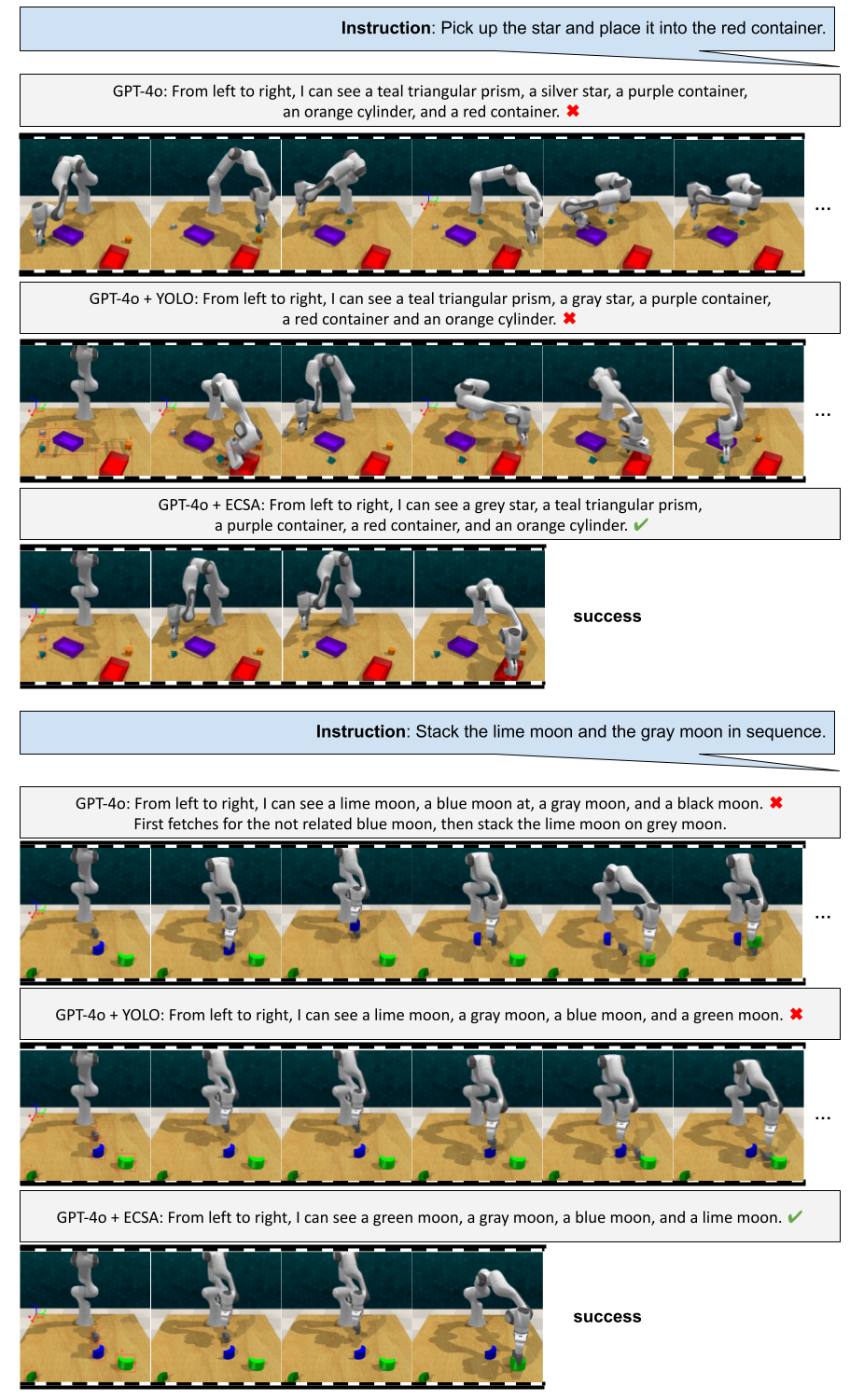}
    \caption{We present two qualitative examples comparing the original MLLM (GPT-4o), its variant augmented with YOLO, and the full \ours pipeline.
Our observations show that \ours improves the precision and quality of scene graph recognition.}
    \label{fig:ebman_qualitative}
\end{figure}

\textbf{Qualitative Studies}
We present two tasks to qualitatively analyze performance on the EB-Manipulation environment.
Each task is solved using three configurations: GPT-4o alone, GPT-4o augmented with YOLO, and GPT-4o augmented with \ours.
Compared to GPT-4o alone, both YOLO and \ours provide additional information that aids task completion.
However, \ours results in more precise recognition than using YOLO alone.

\input{figures/esca_ebman_error_analysis}
\textbf{Error Analysis}
As shown in \figref{fig:eb-man-error-decomposition}, we conducted an error analysis on the EB-Manipulation task.
For each subtask category, we randomly sampled 10 erroneous tasks from both the GPT-4o baseline and GPT-4o augmented with \ours.
The results show that \ours significantly reduces the perception error component.

\newpage

\input{figures/performance-eb-alf}

\subsection{EmbodiedBench: Alfred}

EB-Alfred evaluates an agent's ability to perform household tasks requiring sequential action planning and semantic grounding in realistic indoor environments.
The agent must interpret natural language instructions, navigate through scenes, interact with objects, and complete multi-step tasks such as arranging objects, cleaning, or preparing items.

Tasks are divided into several subcategories: common sense tasks require inferring unstated object properties or goals, complex tasks involve longer action sequences, visual appearance tasks require identifying objects by visual attributes, spatial tasks demand precise spatial reasoning, and long horizon tasks require managing extended sequences of actions. The agent must parse instructions, identify target objects and their states, explore the environment, and execute appropriate action sequences to achieve the specified goals.

\begin{figure}
    \centering
    \includegraphics[height=20cm]{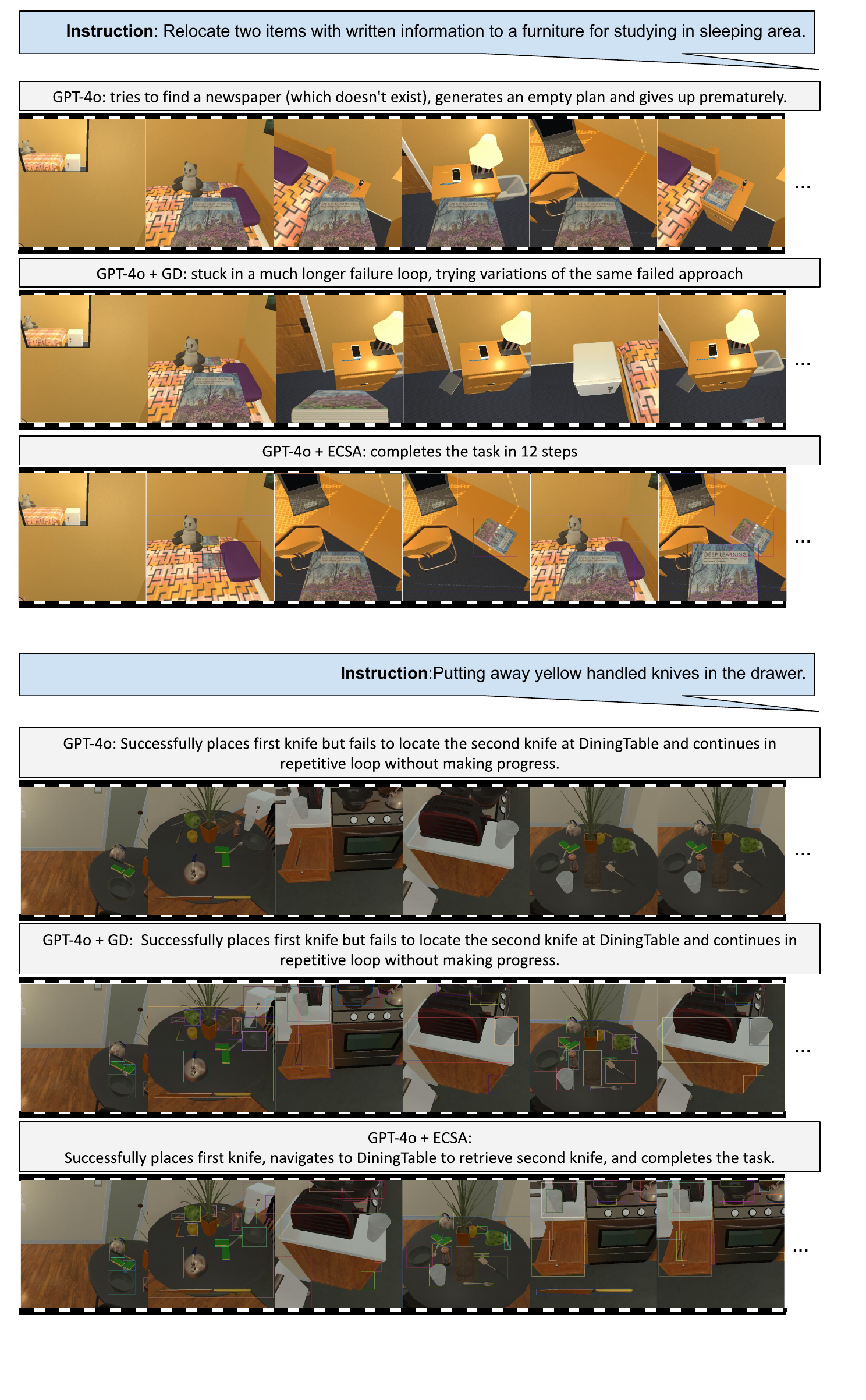}
    \caption{We present two qualitative examples on the Alfred task, comparing the original MLLM (GPT-4o), its variant augmented with Grounding Dino, and the full \ours pipeline.
Our observations show that \ours improves the over all success of the task.}
    \label{fig:ebalfred_qualitative}
\end{figure}

\textbf{Transfer Protocol}:
Similar to EB-Navigation, the overall pipeline of \ours with \ourmodel consists of four main steps: concept extraction, object identification, scene graph generation, and visual summarization and validation.
We begin by using Grounding DINO to extract candidate target objects and related objects based on object names.
Next, \ourmodel predicts categorical labels and spatial relationships for the identified objects, aggregating predictions across detections.
To ensure precision, we retain only the most confident bounding box for each entity in the scene, applying a confidence threshold of $0.3$. Finally, we construct a structured scene description incorporating the identified objects, their positions, and their spatial relationships, and pass this augmented prompt along with the annotated image back to the MLLM to predict subsequent actions.
For Grounding DINO, we use a box threshold of $0.15$ and a text threshold of $0.1$.

\input{figures/esca_ebalf_error_analysis}

\textbf{Qualitative Studies:}
We present two tasks to qualitatively analyze performance on the EB-Alfred environment.
Each task is solved using three configurations: GPT-4o alone, GPT-4o augmented with Grounding DINO, and GPT-4o augmented with \ours.
Compared to GPT-4o alone, both Grounding DINO and \ours provide additional information with the bounding boxes.
However, \ours demonstrates superior spatial reasoning and plan adaptation, successfully completing both tasks where the baseline and Grounding DINO fail due to repetitive loops or premature termination.



\begin{figure}
    \centering
    \includegraphics[height=20cm]{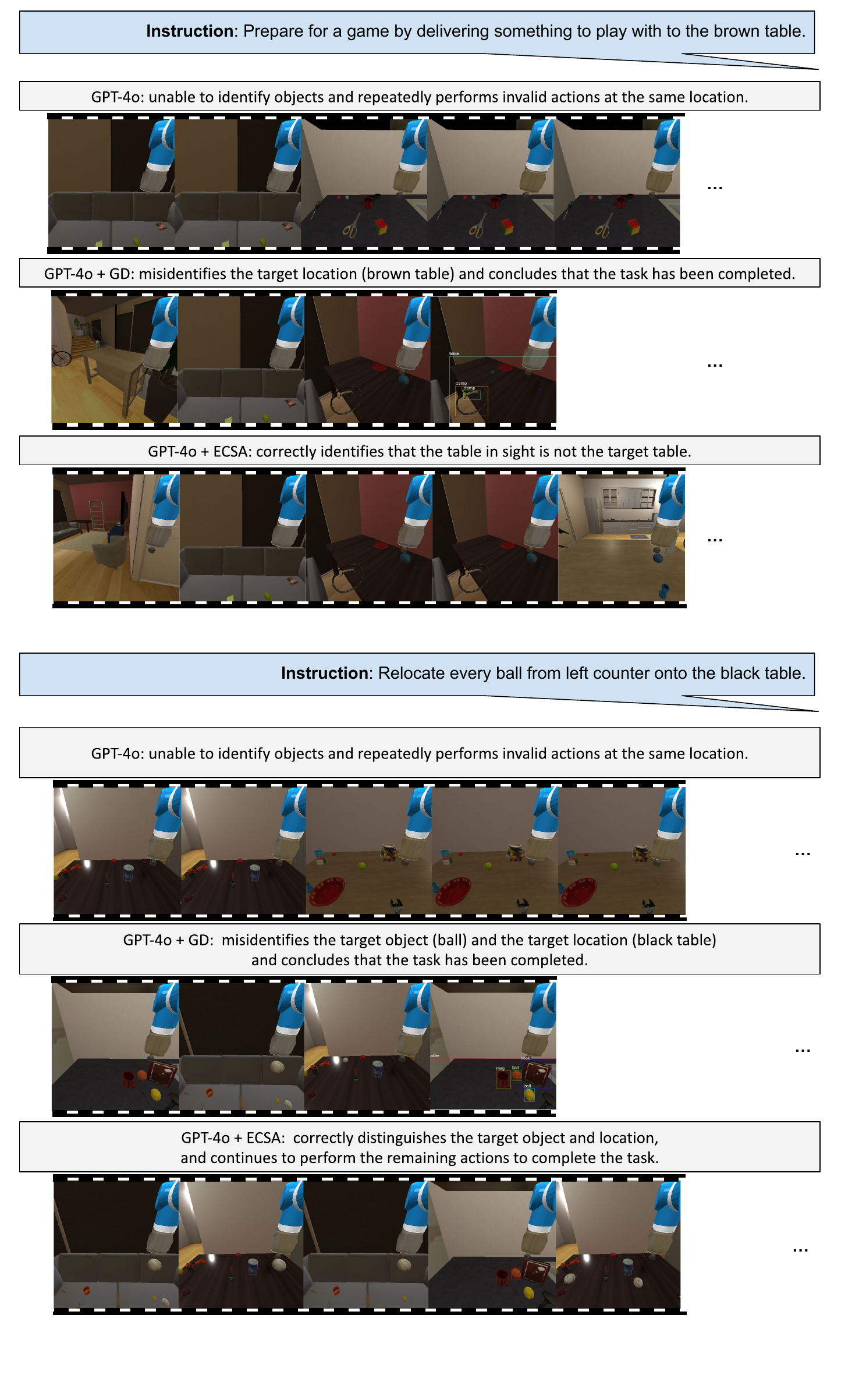}
    \caption{We present two qualitative examples on habitat task comparing the original MLLM (GPT-4o), its variant augmented with Grounding Dino, and the full \ours pipeline.
Our observations show that \ours improves the over all success of the task by providing more detailed attributes in the description.}
    \label{fig:ebhabitat_qualitative}
\end{figure}

\subsection{EmbodiedBench: Habitat}
\input{figures/performance-eb-hab}

EB-Habitat evaluates the agent's high-level task decomposition and planning capabilities.
The agent performs high-level actions to complete the given task in an environment that is interactively navigatable.
Most tasks involve finding an object and moving it to a certain location with slightly different nuances in the formulation of the instruction depending on the subcategory of the task (eg, common sense subtask only specifies what the target object will be used for without naming the specific object itself, long horizon subtask specifies multiple target objects, requiring the agent to perform multiple sequences of actions to complete the task.)
Overall, the agent is required to identify the target object from the instruction, explore the environment to find the object, and correctly move all the objects to the instructed location.

\textbf{Transfer Protocol: Grounding Dino}
Likewise to EB-Navigation, we use the three-step pipeline: concept extraction, object identification, and visual summarization and validation.
The objects of interest are extracted from the instruction as entities with their corresponding current and target states. As EB-Habitat tasks are mostly comprised of moving an object from one place to another, the object states are expressed as a relation between the object and the location in most cases.
The raw object name is input to Grounding DINO and hence used to augment the image with bounding boxes.
The thresholds used for Grounding DINO are: box threshold $=0.15$, text threshold $=0.1$.

\textbf{Transfer Protocol: \ours}
Likewise to EB-Navigation, we use the four-step pipeline, where \ourmodel generates the probability distributions over each entity, keeping only the most confident bounding box in the scene. We apply a confidence threshold of $0.3$, and use a box threshold of $0.15$ and a text threshold of $0.1$ for Grounding DINO.

\input{figures/esca_ebhab_error_analysis}
\textbf{Qualitative Studies}
We present two qualtiative examples to demonstrate the effects of ESCA. Figure~\ref{fig:ebhabitat_qualitative} compares the inference steps of GPT-4o alone, GPT-4o
augmented with Grounding DINO, and GPT-4o augmented with ESCA. In both examples, GPT-4o is unable to identify the object in sight while GPT-4o with Grounding DINO misidentifies the target state as being complete and early terminates the episodes.
On the other hand, ESCA is able to filter out incorrectly identified objects and allows the model to realize that the task is not complete and hence continue its actions to complete the task.


\newpage




\newpage
\subsection{Downstream Task: Scene Graph Generation}
We further analyze the abilities of \ourmodel by applying it to the task of relation tagging of scene graphs. 
In this task, the ground truth bounding boxes of objects are provided, and the objective is to correctly classify their respective object class as well as the binary predicates between them.
To demonstrate \ourmodel on this task, we use the VidVRD \citep{shang2017video} dataset, comprised of 1,000 videos with 35 object categories and 132 predicate categories.

For our evaluation, we finetune two versions of \ourmodel on the VidVRD training set for 75 epochs: the first is the \ourmodel architecture initialized to the original CLIP backbone, denoted as \ourmodel-CLIP; the second is \ourmodel.
The only difference between the two is that the latter has had training from the neurosymbolic learning pipeline denoted in \ref{sec:model-learning}.
We next test these models on 196 videos from the test (4 are ignored for causing an OOM error on a single H100 in both models.)
This test evaluation happens every five epochs; we report the maximum test precision@k and recall@k over all test evaluations for each evaluation in table \ref{table:vidvrd}.
As the table shows, \ourmodel outperforms \ourmodel-CLIP across both metrics for $k=1,5,10$, demonstrating that the proposed neurosymbolic training pipeline provides an advantageous initialization for relational tagging finetuning on VidVRD.

\newcommand{\valep}[2]{#1\textsubscript{\,(#2)}}
\begin{table}[t]
  \centering
  \label{tab:vidvrd_prk}
  \small
  \setlength{\tabcolsep}{5.2pt}
  \begin{tabular}{lcccccc}
    \toprule
     & \multicolumn{2}{c}{$k{=}1$} & \multicolumn{2}{c}{$k{=}5$} & \multicolumn{2}{c}{$k{=}10$} \\
     \cmidrule(lr){2-3} \cmidrule(lr){4-5} \cmidrule(lr){6-7}
    \textbf{Model} & Prec & Rec & Prec & Rec & Prec & Rec \\
    \midrule
    \ourmodel-CLIP & \valep{0.469}{75} & \valep{0.085}{75} & \valep{0.321}{70} & \valep{0.250}{70} & \valep{0.246}{70} & \valep{0.353}{70}\\
    \ourmodel     & \textbf{ \valep{0.495}{45}} & \textbf{ \valep{0.087}{60}} & \textbf{ \valep{0.350}{45}} & \textbf{ \valep{0.270}{75}} & \textbf{ \valep{0.278}{45}} & \textbf{ \valep{0.385}{55}}\\
    \bottomrule
  \end{tabular}
    \caption{Maximum test \textbf{precision}@\textit{k} and \textbf{recall}@\textit{k} on the VidVRD dataset, evaluated every five epochs during 50 epochs of finetuning.  
  Each cell shows the score followed by the epoch at which that score was achieved.}
  \label{table:vidvrd}
\end{table}

\newpage
\subsection{Down-stream Task: Action Recognition}
To evaluate the generalization and finetunability of our model beyond structured scene graph understanding, we consider the task of action recognition, where each video is annotated with a single activity label. We use a combined version of ActivityNet 1.2 and 1.3, which together span 200 unique action classes and approximately 20,000 untrimmed videos across training, validation, and test splits. The action categories range from simple atomic activities such as swimming and rock climbing to more complex, composite tasks like starting a campfire or performing a basketball layup drill.

Similar to Embodied Bench, we adopt a transfer protocol tailored to the action recognition setting. For each second of the video, we generate the top-4 predicted actions along with their confidence scores. We then apply thresholding to discard low-confidence predictions, followed by temporal smoothing to aggregate results into a single continuous segment per video. The final action label is selected based on a weighted combination of its average confidence score and temporal frequency. The output is a single scored segment per video, defined by the merged temporal boundaries of the selected label.

The results, summarized in Table 3, demonstrate the strong fine-tuning capability of ESCA under varying supervision levels. ESCA consistently outperforms CLIP across 0\%, 1\%, and 5\% training subsets, highlighting its superior ability to recognize fine-grained actions. Remarkably, ESCA achieves 92.0\% accuracy using only 5\% of the training data (approximately 800 videos), approaching the performance of InternVL-6B, which is trained on the full dataset. This indicates that ESCA generalizes effectively from limited supervision, whereas InternVL-6B’s advantage stems in part from its exposure to the entire training corpus.
\input{figures/action_recognition}

%% file: figures/performance-eb-nav.tex
\begin{table}[t]
    \centering
    \small
    \begin{tabular}{llrrrrrr}
        \toprule
        \multirow{2}{*}{\textbf{Model}} & \multirow{2}{*}{\textbf{Strategy}} & \multicolumn{5}{c}{\textbf{\textbf{Subset}}} & \multirow{2}{*}{\textbf{Avg}} \\ \cmidrule{3-7}
            &  & Base & Common & Complex & Visual & Long & \\
        \midrule
        \multirow{3}{*}{InternVL-2.5-38B-MPO} 
            & Base       & 55.00 & 60.00 & 51.67 & 40.00 & 30.00 & 47.33 \\
            & + GD            & 60.00 & 61.67 & 56.67 & 48.33 & 11.66 & 47.67 \\
            & + \ours    & 58.33 & 61.67 & 53.33 & 58.33 & 26.66 & \textbf{51.66} \\
        \midrule
        \multirow{3}{*}{Gemini-2.0-flash}
            & Base       & 56.67 & 46.67 & 48.41 & 36.67 & 15.00 & 40.68 \\
            & + GD            & 55.00 & 53.33 & 50.00 & 38.33 & 6.00  & 40.53 \\
            & + \ours    & 56.67 & 41.67 & 43.33 & 46.67 & 21.66 & \textbf{42.00} \\
        \midrule
        \multirow{3}{*}{Qwen2.5-VL-72B-Ins} 
            & Base       & 58.34 & 48.30 & 48.30 & 36.70 & 33.33 & 44.99 \\
            & + GD            & 65.00 & 55.00 &	58.00 &	43.33 &	20.00 &	48.27   \\
            & + \ours    & 60.00  & 48.33   & 60.00   & 46.67   & 31.67   & \textbf{49.33} \\
        \midrule
        \multirow{3}{*}{GPT-4o}
            & Base       & 55.00 & 60.00 & 58.33 & 60.00   & 23.33 & 51.33 \\
            & + GD       & 63.33 &	63.33 &	65.00 &	53.33 &	21.66 & 53.33\\
            & + \ours    & 66.67   & 62.00 & 63.33 & 55.00  & 26.67 & \textbf{54.67} \\
        \bottomrule
    \end{tabular}
    \vspace{5px}
    \caption{
        Detailed performance on EB-Navigation, decomposed by subsets of tasks.
    }
    \label{tab:embodiedbench_nav_detailed}
\end{table}

%% file: figures/eb-manip-algorithm.tex




\begin{algorithm}[H]
\caption{Entropy-Guided Assignment of Entity Names to Objects}
\label{alg:entropy-guided}
\KwIn{
    Objects $O = \{o_1, \dots, o_n\}$;\\
    Entities $E = \{e_1, \dots, e_m\}$ (with possible repetitions);\\
    Assignment scores $\mathcal{S}: O \times E \rightarrow \mathbb{R}$,  $\mathcal{S}(o, e)$ denotes the score of assigning entity $e$ to object $o$
}
\KwOut{Assignment map $A: O \rightarrow E$, assigning one entity name to each object}

Initialize assignment map $A \gets \emptyset$\;

Initialize name availability counter $C: E \rightarrow \mathbb{N}$. how many times each entity name appears in $E$\;

\While{there exists unassigned object $o \in O$}{
    \ForEach{unassigned object $o \in O$}{
        Let $D_o = \left\{ \mathcal{S}(o, e) \,\middle|\, C[e] > 0 \right\}$  \tcp*[r]{Available assignment scores}
    }
    Select object $o^* = \arg\min_{o} \text{Entropy}(D_o)$ \tcp*[r]{Most confident object}
    
    Let $n^* = \arg\max_{e \in E: C[e] > 0} \mathcal{S}(o^*, e)$  \tcp*[r]{Highest scoring available name}
    
    Assign $A[o^*] \gets n^*$ and decrement $C[n^*] \gets C[n^*] - 1$\;
}
\Return $A$\;
\end{algorithm}

%% file: figures/performance-eb-mani.tex
\begin{table}[t]
    \centering
    \small
    \begin{tabular}{llrrrrrr}
        \toprule
        \multirow{2}{*}{\textbf{Model}} & \multirow{2}{*}{\textbf{Strategy}} & \multicolumn{5}{c}{\textbf{\textbf{Subset}}} & \multirow{2}{*}{\textbf{Avg}} \\ \cmidrule{3-7}
            &  & Base & Common & Complex & Visual & Spatial & \\
        \midrule
        \multirow{3}{*}{InternVL-2.5-38B-MPO} & Base & 10.42 & 25.00 & 20.83 & 27.78 & 12.50 & 19.31 \\
         & + YOLO & 14.58 & 14.58 & 25.00 & 19.40 & 22.92 & 19.30 \\
         & + \ours & 31.25 & 21.00 & 14.58 & 27.78 & 27.08 & \textbf{24.30} \\
          \midrule
        \multirow{3}{*}{Gemini-2.0-flash} & Base & 10.42 & 10.42 & 8.33 & 11.11 & 18.75 & 11.81 \\
         & + YOLO & 14.60 & 8.30 & 14.60 & 13.90 & 31.30 & 16.54 \\
         & + \ours & 18.75 & 21.00 & 20.80 & 22.22 & 27.08 & \textbf{21.94} \\  
         \midrule
        \multirow{3}{*}{Qwen2.5-VL-72B-Ins} & Base & 2.08 & 6.25 & 8.33 & 4.16 & 2.78 & 4.72 \\
         & + YOLO & 18.80 & 20.80 & 4.20 & 8.30 & 14.60 & 13.34 \\
         & + \ours & 18.80 & 10.41 & 16.67 & 16.67 & 10.42 & \textbf{14.59} \\
         \midrule
        \multirow{3}{*}{GPT-4o} & Base & 16.67 & 25.00 & 20.83 & 35.41 & 19.44 & 23.47 \\
         & + YOLO & 39.60 & 29.20 & 29.20 & 19.40 & 25.00 & 28.48 \\
         & + \ours & 33.33 & 31.25 & 37.50 & 38.89 & 31.25 & \textbf{34.44} \\
        \bottomrule
    \end{tabular}
    \vspace{5px}
    \caption{
        Detailed performance on EB-Manipulation, decomposed by subsets of tasks.
    }
    \label{tab:grounding-strategies}
\end{table}

%% file: figures/esca_ebman_error_analysis.tex
\begin{figure}
    \footnotesize
    \begin{minipage}{\linewidth}
        \centering
        
        
        \newcommand{\donutchart}[3]{
            \def\radiusfirst{0.5cm}
            \def\radiussecond{1.2cm}
            \def\radiusthird{2cm}
        
            \pgfmathsetlengthmacro\innerradius{\radiussecond}
            \pgfmathsetlengthmacro\outerradius{\radiusthird}
            \pgfmathsetlengthmacro{\centerradius}{(\outerradius + \innerradius)/2}
            \pgfmathsetlengthmacro{\donutcenter}{\innerradius/2}
            
            \pgfmathsetmacro{\totalnum}{0}
            \foreach \value/\colour/\name/\textcolor in {#2} {
                \pgfmathparse{\value+\totalnum}
                \global\let\totalnum=\pgfmathresult
            }
            
            \pgfmathsetmacro{\wheelwidth}{\outerradius-\innerradius}
            \pgfmathsetmacro{\midradius}{(\outerradius+\innerradius)/2}
            
            \begin{scope}[rotate=90]
                \pgfmathsetmacro{\cumnum}{0}
                \foreach \value/\colour/\name/\textcolor in {#2} {
                    \pgfmathsetmacro{\newcumnum}{\cumnum + \value/\totalnum*360}
                    
                    \pgfmathsetmacro{\midangle}{-(\cumnum+\newcumnum)/2}
                    
                    \filldraw[draw=white,fill=\colour] (-\cumnum:\outerradius) arc (-\cumnum:-(\newcumnum):\outerradius) --
                    (-\newcumnum:\innerradius) arc (-\newcumnum:-(\cumnum):\innerradius) -- cycle;
                    
                    \global\let\cumnum=\newcumnum
                }
            \end{scope}
            
            \pgfmathsetlengthmacro\innerradius{\radiusfirst}
            \pgfmathsetlengthmacro\outerradius{\radiussecond}
            \pgfmathsetlengthmacro{\centerradius}{(\outerradius + \innerradius)/2}
            \pgfmathsetlengthmacro{\donutcenter}{\innerradius/2}
            
            \pgfmathsetmacro{\totalnum}{0}
            \foreach \value/\colour/\name/\textcolor in {#3} {
                \pgfmathparse{\value+\totalnum}
                \global\let\totalnum=\pgfmathresult
            }
            
            \pgfmathsetmacro{\wheelwidth}{\outerradius-\innerradius}
            \pgfmathsetmacro{\midradius}{(\outerradius+\innerradius)/2}
            
            \begin{scope}[rotate=90]
                \pgfmathsetmacro{\cumnum}{0}
                \foreach \value/\colour/\name/\textcolor in {#3} {
                    \pgfmathsetmacro{\newcumnum}{\cumnum + \value/\totalnum*360}
                    
                    \pgfmathsetmacro{\midangle}{-(\cumnum+\newcumnum)/2}
                    
                    \filldraw[draw=white,fill=\colour] (-\cumnum:\outerradius) arc (-\cumnum:-(\newcumnum):\outerradius) --
                    (-\newcumnum:\innerradius) arc (-\newcumnum:-(\cumnum):\innerradius) -- cycle;
                    
                    \global\let\cumnum=\newcumnum
                }
            \end{scope}
            
            \pgfmathsetlengthmacro\innerradius{\radiussecond}
            \pgfmathsetlengthmacro\outerradius{\radiusthird}
            \pgfmathsetlengthmacro{\centerradius}{(\outerradius + \innerradius)/2}
            \pgfmathsetlengthmacro{\donutcenter}{\innerradius/2}
            
            \pgfmathsetmacro{\totalnum}{0}
            \foreach \value/\colour/\name/\textcolor in {#2} {
                \pgfmathparse{\value+\totalnum}
                \global\let\totalnum=\pgfmathresult
            }
            
            \pgfmathsetmacro{\wheelwidth}{\outerradius-\innerradius}
            \pgfmathsetmacro{\midradius}{(\outerradius+\innerradius)/2}
            
            \begin{scope}[rotate=90]
                \pgfmathsetmacro{\cumnum}{0}
                \foreach \value/\colour/\name/\textcolor in {#2} {
                    \pgfmathsetmacro{\newcumnum}{\cumnum + \value/\totalnum*360}
                    
                    \pgfmathsetmacro{\midangle}{-(\cumnum+\newcumnum)/2}
                    
                    \draw[fill=none] node [font=\tiny, color=\textcolor] at (\midangle:{\innerradius+\wheelwidth/2}) {\name};
                    
                    \global\let\cumnum=\newcumnum
                }
            \end{scope}
            
            \pgfmathsetlengthmacro\innerradius{\radiusfirst}
            \pgfmathsetlengthmacro\outerradius{\radiussecond}
            \pgfmathsetlengthmacro{\centerradius}{(\outerradius + \innerradius)/2}
            \pgfmathsetlengthmacro{\donutcenter}{\innerradius/2}
            
            \pgfmathsetmacro{\totalnum}{0}
            \foreach \value/\colour/\name/\textcolor in {#3} {
                \pgfmathparse{\value+\totalnum}
                \global\let\totalnum=\pgfmathresult
            }
            
            \pgfmathsetmacro{\wheelwidth}{\outerradius-\innerradius}
            \pgfmathsetmacro{\midradius}{(\outerradius+\innerradius)/2}
            
            \begin{scope}[rotate=90]
                \pgfmathsetmacro{\cumnum}{0}
                \foreach \value/\colour/\name/\textcolor in {#3} {
                    \pgfmathsetmacro{\newcumnum}{\cumnum + \value/\totalnum*360}
                    
                    \pgfmathsetmacro{\midangle}{-(\cumnum+\newcumnum)/2}
                    
                    \draw[fill=none] node [font=\scriptsize, color=\textcolor, align=center] at (\midangle:{\innerradius+\wheelwidth/2}) {\name};
                    
                    \global\let\cumnum=\newcumnum
                }
            \end{scope}
        }%
        \begin{tikzpicture}
            \donutchart{+ ESCA}{
                12/deepseek2!30!white/WR/deepseek2!60!black,
                12/deepseek2!30!white/SU/deepseek2!60!black, 
                14/llamaOrange!30!white/RE/llamaOrange!60!black, 
                62/gpt1!30!white/IA/gpt1!60!black%
            }{
                24/deepseek2!60!white/Perc.\\(24\%)/deepseek2!40!black,
                14/llamaOrange!60!white/Reas.\\(14\%)/llamaOrange!40!black,
                62/gpt1!60!white/Plan.\\(62\%)/gpt1!40!black%
            }
            \draw node [align=center] at (-3cm,0) {GPT-4o\\+ ESCA};
        \end{tikzpicture}
        \begin{tikzpicture}
            \donutchart{Base}{
                14/deepseek2!30!white/SU/deepseek2!60!black, 
                26/deepseek2!30!white/WR/deepseek2!60!black,
                10/llamaOrange!30!white/
                /llamaOrange!60!black, 
                50/gpt1!30!white/IA/gpt1!60!black%
            }{
                40/deepseek2!60!white/Perc.\\(40\%)/deepseek2!40!black,
                10/llamaOrange!60!white/Reas.\\(10\%)/llamaOrange!40!black,
                50/gpt1!60!white/Plan. (50\%)/gpt1!40!black%
            }
            \draw node [align=center] at (-3cm,0) {GPT-4o};
        \end{tikzpicture}
        
        \vspace{-3px}
        
        \captionof{figure}{
            Error decomposition\protect\footnotemark~of GPT-4o with and without \ours, based on manual inspection of $50$ EB-Manipulation tasks, with $10$ randomly sampled from each subtasks.
            As we can see, \ours has reduced perception error by a large margin.
        }
        \label{fig:eb-man-error-decomposition}
    \end{minipage}
\end{figure}
\footnotetext{
    The three top-level error types are \underline{Perc}eption, \underline{Reas}oning, and \underline{Plan}ning. 
    The second-level errors are \underline{Ha}llucination, \underline{W}rong \underline{R}ecognition, \underline{S}patial \underline{U}nderstanding, \underline{S}patial \underline{R}easoning, \underline{R}eflection \underline{E}rror, \underline{I}naccurate \underline{A}ction, and \underline{Co}llision.
    For clarity, the figure uses these acronyms to label the different error types.
}

%% file: figures/performance-eb-alf.tex
\begin{table}[t]
    \centering
    \small
    \begin{tabular}{llrrrrrrr}
        \toprule
        \multirow{2}{*}{\textbf{Model}} & \multirow{2}{*}{\textbf{Strategy}} & \multicolumn{6}{c}{\textbf{Subset}} & \multirow{2}{*}{\textbf{Avg}} \\ \cmidrule{3-8}
            &  & Base & Common & Complex & Visual & Spatial & Long & \\
        \midrule
        \multirow{3}{*}{InternVL-2.5-38B-MPO} 
            & Base      & 32.00 & 20.00 & 20.00 &  18.00 & 20.00 & 50.00 & 26.67 \\
            & + GD      & 28.00 & 36.00 & 34.00 & 32.00 & 30.00 & 47.83 & 34.64 \\
            & + \ours   & 43.00 & 32.00 & 46.00 & 36.00 & 24.00 & 50.00 & \textbf{38.50} \\
        \midrule
        \multirow{3}{*}{Gemini-2.0-flash} 
            & Base      & 68.00 & 58.00 & 52.00 & 46.00 & 42.00 & 52.00 & 53.00 \\
            & + GD      & 68.00 & 58.00 & 52.00 & 46.00 & 42.00 & 56.00 & 53.67 \\
            & + \ours   & 70.00 & 54.00 & 56.00 & 54.00 & 44.00 & 50.00 & \textbf{54.67} \\
        \midrule
        \multirow{3}{*}{Qwen2.5-VL-72B-Ins} 
            & Base      & 50.00 & 42.00 & 42.00 & 36.00 & 34.00 & 34.00 & 39.67 \\
            & + GD      & 52.00 & 38.00 & 50.00 & 46.00 & 40.00 & 42.00 & \textbf{44.67} \\
            & + \ours   & 46.00 & 36.00 & 54.00 & 47.00 & 37.00 & 30.00 & 41.67 \\
        \midrule
        \multirow{3}{*}{GPT-4o} 
            & Base      & 64.00 & 48.00 & 62.00 & 46.00 & 50.00 & 54.00 & 54.00 \\
            & + GD      & 52.00 & 46.00 & 62.00 & 48.00 & 50.00 & 56.00 & 52.33 \\
            & + \ours   & 62.00 & 56.00 & 54.00 & 53.00 & 52.00 & 50.00 & \textbf{54.50} \\
        \bottomrule
    \end{tabular}
    \vspace{5px}
    \caption{Detailed performance on EB-Alfred, decomposed by subsets of tasks.}
    \label{tab:grounding-strategies-alf}
\end{table}

%% file: figures/esca_ebalf_error_analysis.tex
\begin{figure}
    \footnotesize
    \begin{minipage}{\linewidth}
        \centering
        
        
        \newcommand{\donutchart}[3]{
            \def\radiusfirst{0.5cm}
            \def\radiussecond{1.2cm}
            \def\radiusthird{2cm}
        
            \pgfmathsetlengthmacro\innerradius{\radiussecond}
            \pgfmathsetlengthmacro\outerradius{\radiusthird}
            \pgfmathsetlengthmacro{\centerradius}{(\outerradius + \innerradius)/2}
            \pgfmathsetlengthmacro{\donutcenter}{\innerradius/2}
            
            \pgfmathsetmacro{\totalnum}{0}
            \foreach \value/\colour/\name/\textcolor in {#2} {
                \pgfmathparse{\value+\totalnum}
                \global\let\totalnum=\pgfmathresult
            }
            
            \pgfmathsetmacro{\wheelwidth}{\outerradius-\innerradius}
            \pgfmathsetmacro{\midradius}{(\outerradius+\innerradius)/2}
            
            \begin{scope}[rotate=90]
                \pgfmathsetmacro{\cumnum}{0}
                \foreach \value/\colour/\name/\textcolor in {#2} {
                    \pgfmathsetmacro{\newcumnum}{\cumnum + \value/\totalnum*360}
                    
                    \pgfmathsetmacro{\midangle}{-(\cumnum+\newcumnum)/2}
                    
                    \filldraw[draw=white,fill=\colour] (-\cumnum:\outerradius) arc (-\cumnum:-(\newcumnum):\outerradius) --
                    (-\newcumnum:\innerradius) arc (-\newcumnum:-(\cumnum):\innerradius) -- cycle;
                    
                    \global\let\cumnum=\newcumnum
                }
            \end{scope}
            
            \pgfmathsetlengthmacro\innerradius{\radiusfirst}
            \pgfmathsetlengthmacro\outerradius{\radiussecond}
            \pgfmathsetlengthmacro{\centerradius}{(\outerradius + \innerradius)/2}
            \pgfmathsetlengthmacro{\donutcenter}{\innerradius/2}
            
            \pgfmathsetmacro{\totalnum}{0}
            \foreach \value/\colour/\name/\textcolor in {#3} {
                \pgfmathparse{\value+\totalnum}
                \global\let\totalnum=\pgfmathresult
            }
            
            \pgfmathsetmacro{\wheelwidth}{\outerradius-\innerradius}
            \pgfmathsetmacro{\midradius}{(\outerradius+\innerradius)/2}
            
            \begin{scope}[rotate=90]
                \pgfmathsetmacro{\cumnum}{0}
                \foreach \value/\colour/\name/\textcolor in {#3} {
                    \pgfmathsetmacro{\newcumnum}{\cumnum + \value/\totalnum*360}
                    
                    \pgfmathsetmacro{\midangle}{-(\cumnum+\newcumnum)/2}
                    
                    \filldraw[draw=white,fill=\colour] (-\cumnum:\outerradius) arc (-\cumnum:-(\newcumnum):\outerradius) --
                    (-\newcumnum:\innerradius) arc (-\newcumnum:-(\cumnum):\innerradius) -- cycle;
                    
                    \global\let\cumnum=\newcumnum
                }
            \end{scope}
            
            \pgfmathsetlengthmacro\innerradius{\radiussecond}
            \pgfmathsetlengthmacro\outerradius{\radiusthird}
            \pgfmathsetlengthmacro{\centerradius}{(\outerradius + \innerradius)/2}
            \pgfmathsetlengthmacro{\donutcenter}{\innerradius/2}
            
            \pgfmathsetmacro{\totalnum}{0}
            \foreach \value/\colour/\name/\textcolor in {#2} {
                \pgfmathparse{\value+\totalnum}
                \global\let\totalnum=\pgfmathresult
            }
            
            \pgfmathsetmacro{\wheelwidth}{\outerradius-\innerradius}
            \pgfmathsetmacro{\midradius}{(\outerradius+\innerradius)/2}
            
            \begin{scope}[rotate=90]
                \pgfmathsetmacro{\cumnum}{0}
                \foreach \value/\colour/\name/\textcolor in {#2} {
                    \pgfmathsetmacro{\newcumnum}{\cumnum + \value/\totalnum*360}
                    
                    \pgfmathsetmacro{\midangle}{-(\cumnum+\newcumnum)/2}
                    
                    \draw[fill=none] node [font=\tiny, color=\textcolor] at (\midangle:{\innerradius+\wheelwidth/2}) {\name};
                    
                    \global\let\cumnum=\newcumnum
                }
            \end{scope}
            
            \pgfmathsetlengthmacro\innerradius{\radiusfirst}
            \pgfmathsetlengthmacro\outerradius{\radiussecond}
            \pgfmathsetlengthmacro{\centerradius}{(\outerradius + \innerradius)/2}
            \pgfmathsetlengthmacro{\donutcenter}{\innerradius/2}
            
            \pgfmathsetmacro{\totalnum}{0}
            \foreach \value/\colour/\name/\textcolor in {#3} {
                \pgfmathparse{\value+\totalnum}
                \global\let\totalnum=\pgfmathresult
            }
            
            \pgfmathsetmacro{\wheelwidth}{\outerradius-\innerradius}
            \pgfmathsetmacro{\midradius}{(\outerradius+\innerradius)/2}
            
            \begin{scope}[rotate=90]
                \pgfmathsetmacro{\cumnum}{0}
                \foreach \value/\colour/\name/\textcolor in {#3} {
                    \pgfmathsetmacro{\newcumnum}{\cumnum + \value/\totalnum*360}
                    
                    \pgfmathsetmacro{\midangle}{-(\cumnum+\newcumnum)/2}
                    
                    \draw[fill=none] node [font=\scriptsize, color=\textcolor, align=center] at (\midangle:{\innerradius+\wheelwidth/2}) {\name};
                    
                    \global\let\cumnum=\newcumnum
                }
            \end{scope}
        }%
        
        \begin{tikzpicture}
            \donutchart{+ ESCA}{
                4/deepseek2!30!white/WR/deepseek2!60!black,
                2/llamaOrange!30!white/RE/llamaOrange!60!black, 
                15/gpt1!30!white/IA/gpt1!60!black%
            }{
                4/deepseek2!60!white/Perc.\\(19\%)/deepseek2!40!black,
                2/llamaOrange!60!white/Reas.\\(10\%)/llamaOrange!40!black,
                15/gpt1!60!white/Plan.\\(71\%)/gpt1!40!black%
            }
            \draw node [align=center] at (-3cm,0) {GPT-4o\\+ ESCA};
        \end{tikzpicture}
        \hspace{1cm}
        \begin{tikzpicture}
            \donutchart{Base}{
                1/deepseek2!30!white/Ha/deepseek2!60!black, 
                6/deepseek2!30!white/WR/deepseek2!60!black,
                2/llamaOrange!30!white/SR/llamaOrange!60!black, 
                5/llamaOrange!30!white/RE/llamaOrange!60!black, 
                10/gpt1!30!white/IA/gpt1!60!black%
            }{
                7/deepseek2!60!white/Perc.\\(29\%)/deepseek2!40!black,
                7/llamaOrange!60!white/Reas.\\(29\%)/llamaOrange!40!black,
                10/gpt1!60!white/Plan.\\(42\%)/gpt1!40!black%
            }
            \draw node [align=center] at (-3cm,0) {GPT-4o};
        \end{tikzpicture}
        
        \vspace{0.3cm}
        
        \captionof{figure}{
            Error decomposition\protect\footnotemark~of GPT-4o with and without \ours, based on manual inspection of $60$ EB-Alfred tasks, with $10$ randomly sampled from each subtask.
            \ours reduces perception errors (Ha, WR) and reasoning errors (RE, SR), shifting failures to planning stages where action sequencing can be refined.
        }
        \label{fig:eb-nav-error-decomposition}
    \end{minipage}
\end{figure}

\footnotetext{
    The three top-level error types are \underline{Perc}eption, \underline{Reas}oning, and \underline{Plan}ning. 
    The second-level errors are \underline{Ha}llucination, \underline{W}rong \underline{R}ecognition, \underline{S}patial \underline{U}nderstanding, \underline{S}patial \underline{R}easoning, \underline{R}eflection \underline{E}rror, and \underline{I}naccurate \underline{A}ction.
    For clarity, the figure uses these acronyms to label the different error types.
}

%% file: figures/performance-eb-hab.tex
\begin{table}[t]
    \centering
    \small
    \begin{tabular}{llrrrrrrr}
        \toprule
        \multirow{2}{*}{\textbf{Model}} & \multirow{2}{*}{\textbf{Strategy}} & \multicolumn{6}{c}{\textbf{Subset}} & \multirow{2}{*}{\textbf{Avg}} \\ \cmidrule{3-8}
            &  & Base & Common & Complex & Visual & Spatial & Long & \\
        \midrule
        \multirow{3}{*}{InternVL-2.5-38B-MPO} 
            & Base      & 92.00 & 24.00 & 64.00 & 46.00 & 30.00 & 32.00 & 48.00 \\
            & + GD      & 84.00 & 36.00 & 66.00 & 52.00 & 32.00 & 40.00 & 51.67 \\
            & + \ours   & 94.00 & 46.00 & 62.00 & 54.00 & 34.00 & 40.00 & \textbf{55.00} \\
        \midrule
        \multirow{3}{*}{Gemini-2.0-flash} 
            & Base      & 66.00 & 20.00 & 32.00 & 32.00 & 20.04 & 10.20 & 30.04 \\
            & + GD      & 68.00 & 18.36 & 32.00 & 24.00 & 20.41 & 10.00 & 28.80 \\
            & + \ours   & 74.00 & 22.00 & 35.14 & 36.00 & 22.00 & 14.00 & \textbf{33.86} \\
        \midrule
        \multirow{3}{*}{Qwen2.5-VL-72B-Ins} 
            & Base      & 58.00 & 18.00 & 42.00 & 40.00 & 24.00 & 18.00 & 33.33 \\
            & + GD      & 52.00 & 50.00 & 68.00 & 70.00 & 21.00 & 36.00 & 49.50 \\
            & + \ours   & 86.00 & 50.00 & 72.00 & 60.00 & 36.00 & 34.00 & \textbf{56.33} \\
        \midrule
        \multirow{3}{*}{GPT-4o} 
            & Base      & 92.00 & 46.00 & 52.00 & 66.00 & 32.65 & 59.18 & 57.97 \\
            & + GD      & 86.00 & 58.00 & 62.00 & 76.00 & 32.00 & 34.00 & 58.00 \\
            & + \ours   & 88.00 & 62.00 & 58.00 & 72.00 & 32.00 & 48.00 & \textbf{60.00} \\
        \bottomrule
    \end{tabular}
    \vspace{5px}
    \caption{Detailed performance on EB-Habitat, decomposed by subsets of tasks.}
    \label{tab:grounding-strategies-hab}
\end{table}

%% file: figures/esca_ebhab_error_analysis.tex
\begin{figure}
    \footnotesize
    \begin{minipage}{\linewidth}
        \centering
        
        
        \newcommand{\donutchart}[3]{
            \def\radiusfirst{0.5cm}
            \def\radiussecond{1.2cm}
            \def\radiusthird{2cm}
        
            \pgfmathsetlengthmacro\innerradius{\radiussecond}
            \pgfmathsetlengthmacro\outerradius{\radiusthird}
            \pgfmathsetlengthmacro{\centerradius}{(\outerradius + \innerradius)/2}
            \pgfmathsetlengthmacro{\donutcenter}{\innerradius/2}
            
            \pgfmathsetmacro{\totalnum}{0}
            \foreach \value/\colour/\name/\textcolor in {#2} {
                \pgfmathparse{\value+\totalnum}
                \global\let\totalnum=\pgfmathresult
            }
            
            \pgfmathsetmacro{\wheelwidth}{\outerradius-\innerradius}
            \pgfmathsetmacro{\midradius}{(\outerradius+\innerradius)/2}
            
            \begin{scope}[rotate=90]
                \pgfmathsetmacro{\cumnum}{0}
                \foreach \value/\colour/\name/\textcolor in {#2} {
                    \pgfmathsetmacro{\newcumnum}{\cumnum + \value/\totalnum*360}
                    
                    \pgfmathsetmacro{\midangle}{-(\cumnum+\newcumnum)/2}
                    
                    \filldraw[draw=white,fill=\colour] (-\cumnum:\outerradius) arc (-\cumnum:-(\newcumnum):\outerradius) --
                    (-\newcumnum:\innerradius) arc (-\newcumnum:-(\cumnum):\innerradius) -- cycle;
                    
                    \global\let\cumnum=\newcumnum
                }
            \end{scope}
            
            \pgfmathsetlengthmacro\innerradius{\radiusfirst}
            \pgfmathsetlengthmacro\outerradius{\radiussecond}
            \pgfmathsetlengthmacro{\centerradius}{(\outerradius + \innerradius)/2}
            \pgfmathsetlengthmacro{\donutcenter}{\innerradius/2}
            
            \pgfmathsetmacro{\totalnum}{0}
            \foreach \value/\colour/\name/\textcolor in {#3} {
                \pgfmathparse{\value+\totalnum}
                \global\let\totalnum=\pgfmathresult
            }
            
            \pgfmathsetmacro{\wheelwidth}{\outerradius-\innerradius}
            \pgfmathsetmacro{\midradius}{(\outerradius+\innerradius)/2}
            
            \begin{scope}[rotate=90]
                \pgfmathsetmacro{\cumnum}{0}
                \foreach \value/\colour/\name/\textcolor in {#3} {
                    \pgfmathsetmacro{\newcumnum}{\cumnum + \value/\totalnum*360}
                    
                    \pgfmathsetmacro{\midangle}{-(\cumnum+\newcumnum)/2}
                    
                    \filldraw[draw=white,fill=\colour] (-\cumnum:\outerradius) arc (-\cumnum:-(\newcumnum):\outerradius) --
                    (-\newcumnum:\innerradius) arc (-\newcumnum:-(\cumnum):\innerradius) -- cycle;
                    
                    \global\let\cumnum=\newcumnum
                }
            \end{scope}
            
            \pgfmathsetlengthmacro\innerradius{\radiussecond}
            \pgfmathsetlengthmacro\outerradius{\radiusthird}
            \pgfmathsetlengthmacro{\centerradius}{(\outerradius + \innerradius)/2}
            \pgfmathsetlengthmacro{\donutcenter}{\innerradius/2}
            
            \pgfmathsetmacro{\totalnum}{0}
            \foreach \value/\colour/\name/\textcolor in {#2} {
                \pgfmathparse{\value+\totalnum}
                \global\let\totalnum=\pgfmathresult
            }
            
            \pgfmathsetmacro{\wheelwidth}{\outerradius-\innerradius}
            \pgfmathsetmacro{\midradius}{(\outerradius+\innerradius)/2}
            
            \begin{scope}[rotate=90]
                \pgfmathsetmacro{\cumnum}{0}
                \foreach \value/\colour/\name/\textcolor in {#2} {
                    \pgfmathsetmacro{\newcumnum}{\cumnum + \value/\totalnum*360}
                    
                    \pgfmathsetmacro{\midangle}{-(\cumnum+\newcumnum)/2}
                    
                    \draw[fill=none] node [font=\tiny, color=\textcolor] at (\midangle:{\innerradius+\wheelwidth/2}) {\name};
                    
                    \global\let\cumnum=\newcumnum
                }
            \end{scope}
            
            \pgfmathsetlengthmacro\innerradius{\radiusfirst}
            \pgfmathsetlengthmacro\outerradius{\radiussecond}
            \pgfmathsetlengthmacro{\centerradius}{(\outerradius + \innerradius)/2}
            \pgfmathsetlengthmacro{\donutcenter}{\innerradius/2}
            
            \pgfmathsetmacro{\totalnum}{0}
            \foreach \value/\colour/\name/\textcolor in {#3} {
                \pgfmathparse{\value+\totalnum}
                \global\let\totalnum=\pgfmathresult
            }
            
            \pgfmathsetmacro{\wheelwidth}{\outerradius-\innerradius}
            \pgfmathsetmacro{\midradius}{(\outerradius+\innerradius)/2}
            
            \begin{scope}[rotate=90]
                \pgfmathsetmacro{\cumnum}{0}
                \foreach \value/\colour/\name/\textcolor in {#3} {
                    \pgfmathsetmacro{\newcumnum}{\cumnum + \value/\totalnum*360}
                    
                    \pgfmathsetmacro{\midangle}{-(\cumnum+\newcumnum)/2}
                    
                    \draw[fill=none] node [font=\scriptsize, color=\textcolor, align=center] at (\midangle:{\innerradius+\wheelwidth/2}) {\name};
                    
                    \global\let\cumnum=\newcumnum
                }
            \end{scope}
        }%
        \begin{tikzpicture}
            \donutchart{+ ESCA}{
                8/llamaOrange!30!white/IE/llamaOrange!60!black,
                8/llamaOrange!30!white/RE/llamaOrange!60!black, 
                9/deepseek2!30!white/WR/deepseek2!60!black, 
                2/gpt1!30!white/IA/gpt1!60!black%
            }{
                16/llamaOrange!60!white/Reas.\\(60\%)/llamaOrange!40!black,
                9/deepseek2!60!white/Perc.\\(33\%)/deepseek2!40!black,
                2/gpt1!60!white/Plan.\\(7\%)/gpt1!40!black%
            }
            \draw node [align=center] at (-3cm,0) {GPT-4o\\+ ESCA};
        \end{tikzpicture}
        \begin{tikzpicture}
            \donutchart{Base}{
                5/llamaOrange!30!white/IE/llamaOrange!60!black, 
                8/llamaOrange!30!white/RE/llamaOrange!60!black, 
                10/deepseek2!30!white/WR/deepseek2!60!black,
                3/gpt1!30!white/IA/gpt1!60!black%
            }{
                13/llamaOrange!60!white/Reas.\\(50\%)/llamaOrange!40!black,
                10/deepseek2!60!white/Perc.\\(38\%)/deepseek2!40!black,
                3/gpt1!60!white/Plan. (12\%)/gpt1!40!black%
            }
            \draw node [align=center] at (-3cm,0) {GPT-4o};
        \end{tikzpicture}
        
        \vspace{-3px}
        
        \captionof{figure}{
            Error decomposition\protect\footnotemark~of GPT-4o with and without \ours, based on manual inspection of $50$ EB-Habitat tasks, with $10$ randomly sampled from each subtasks.
        }
        \label{fig:eb-man-error-decomposition}
    \end{minipage}
\end{figure}
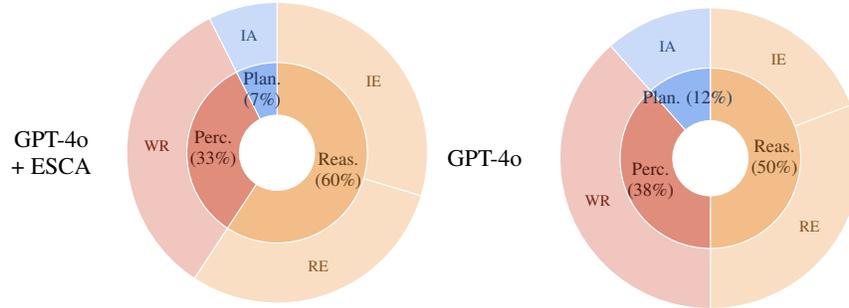
\footnotetext{
    The three top-level error types are \underline{Perc}eption, \underline{Reas}oning, and \underline{Plan}ning. 
    The second-level errors are \underline{Ha}llucination, \underline{W}rong \underline{R}ecognition, \underline{S}patial \underline{U}nderstanding, \underline{S}patial \underline{R}easoning, \underline{R}eflection \underline{E}rror, \underline{I}naccurate \underline{A}ction, and \underline{Co}llision.
    For clarity, the figure uses these acronyms to label the different error types.
}

%% file: figures/action_recognition.tex
\begin{table}[t]
\centering
\small
\begin{tabular}{llr}
\toprule
\textbf{Category} & \textbf{Model} & \textbf{ActivityNet Accuracy (\%)} \\
\midrule
\multirow{5}{*}{Zero-shot} 
 & \ourmodel & 76.34 \\
 & CLIP  & 74.37 \\
 & BIKE & 80.00 \\
 & Text4vis & 77.40 \\
 & ResT & 26.30 \\
 & E2E & 20.00 \\
\midrule
\multirow{2}{*}{Few-shot (1\%)}
 & \ourmodel & 80.10 \\
 & CLIP  & 78.79 \\
\midrule
\multirow{2}{*}{Few-shot (5\%)}
 & \ourmodel & 86.05 \\
 & CLIP & 80.02 \\
\midrule
Fully finetuned & InternVL-6B & 95.90 \\
\bottomrule
\end{tabular}
\caption{ActivityNet accuracy for zero-shot, few-shot, and fully finetuned models. Zero-shot includes both external baselines and our models evaluated without training.}
\label{tab:activitynet_comparison}
\end{table}